\documentclass{article}

    \usepackage[final,nonatbib]{neurips_2020}

\usepackage[utf8]{inputenc} %
\usepackage[T1]{fontenc}    %
\usepackage{url}            %
\usepackage{booktabs}       %
\usepackage{amsfonts}       %
\usepackage{nicefrac}       %
\usepackage{microtype}      %

\usepackage[numbers]{natbib}

\usepackage{amsmath,amsfonts,bm}

\def\eqref#1{equation~\ref{#1}}

\def\1{\bm{1}}

\def\vb{{\bm{b}}}

\def\vg{{\bm{g}}}
\def\vh{{\bm{h}}}

\def\vq{{\bm{q}}}

\def\vu{{\bm{u}}}
\def\vv{{\bm{v}}}
\def\vw{{\bm{w}}}
\def\vx{{\bm{x}}}

\def\vz{{\bm{z}}}

\def\mA{{\bm{A}}}
\def\mB{{\bm{B}}}

\def\mI{{\bm{I}}}

\def\mU{{\bm{U}}}

\def\mW{{\bm{W}}}
\def\mX{{\bm{X}}}

\DeclareMathAlphabet{\mathsfit}{\encodingdefault}{\sfdefault}{m}{sl}
\SetMathAlphabet{\mathsfit}{bold}{\encodingdefault}{\sfdefault}{bx}{n}

\newcommand{\E}{\mathbb{E}}

\newcommand{\Var}{\mathrm{VAR}}

\newcommand{\vectorize}{\mathrm{vec}}

\DeclareMathOperator*{\argmin}{arg\,min}

\def\comma{{ \text{ ,} }}
\def\period{{ \text{ .} }}

\DeclareMathOperator{\diag}{diag}

\def\ttranspose{{t \text{ } \transpose}}
\def\vxr{{\vx_r}}

\def\kt{{\mathbf{k}_t}}
\def\Kt{{\mathbf{K}_t}}
\def\Gt{{\mathbf{G}_t}}

\def\KtT{{\mathbf{K}_t^\transpose}}
\def\GtT{{\mathbf{G}_t^\transpose}}
\def\GtsT{{\mathbf{G}_{t_s}^\transpose}}

\def\It{{\mathbf{I}_t}}
\def\Lt{{\mathbf{L}_t}}

\def\ku{{\tilde{\mathbf{k}}}}
\def\kut{{\tilde{\mathbf{k}}_t}}
\def\Kut{{\tilde{\mathbf{K}}_t}}
\def\Gut{{\tilde{\mathbf{G}}_t}}

\def\Iv{{\tilde{\mathbf{I}}}}
\def\Ivt{{\tilde{\mathbf{I}}_t}}
\def\Lvt{{\tilde{\mathbf{L}}_t}}
\def\Hvt{{\tilde{\mathbf{H}}_t}}

\def\QuuC{{\tilde{Q}_{\vu \vu}}}

\def\QuutC{{\tilde{Q}^t_{\vu \vu}}}
\def\QvvtC{{\tilde{Q}^t_{\vv \vv}}}
\def\QuuCInv{{\tilde{Q}^{-1}_{\vu \vu}}}
\def\QvvCInv{{\tilde{Q}^{-1}_{\vv \vv}}}
\def\QA{{Q_{\vu \vx_r}}}
\def\QB{{Q_{\vv \vx}}}
\def\QAt{{Q^t_{\vu \vx_r}}}
\def\QBt{{Q^t_{\vv \vx}}}

\def\Axx{{\E[\vx_{\vu}\vx_{\vu}^\transpose]}}
\def\Axy{{\E[\vx_{\vu}\vx_{\vv}^\transpose]}}
\def\Ayy{{\E[\vx_{\vv}\vx_{\vv}^\transpose]}}

\def\Bxx{{\E[\vg_{\vu}\vg_{\vu}^\transpose]}}
\def\Bxy{{\E[\vg_{\vu}\vg_{\vv}^\transpose]}}
\def\Byy{{\E[\vg_{\vv}\vg_{\vv}^\transpose]}}

\def\Auu{{A_{\vu\vu}}}
\def\Auv{{A_{\vu\vv}}}
\def\Avu{{A_{\vv\vu}}}
\def\Avv{{A_{\vv\vv}}}
\def\Buu{{B_{\vu\vu}}}
\def\Buv{{B_{\vu\vv}}}
\def\Bvu{{B_{\vv\vu}}}
\def\Bvv{{B_{\vv\vv}}}
\def\AuvT{{A^\transpose_{\vu\vv}}}
\def\BuvT{{B^\transpose_{\vu\vv}}}

\def\AuuInv{{A^{-1}_{\vu\vu}}}
\def\AuuInvT{{A^{-\transpose}_{\vu\vu}}}
\def\AvvInv{{A^{-1}_{\vv\vv}}}
\def\BuuInv{{B^{-1}_{\vu\vu}}}
\def\BvvInv{{B^{-1}_{\vv\vv}}}
\def\AuuC{{\tilde{A}_{\vu\vu}}}
\def\AvvC{{\tilde{A}_{\vv\vv}}}
\def\BuuC{{\tilde{B}_{\vu\vu}}}
\def\BvvC{{\tilde{B}_{\vv\vv}}}

\def\AuuCInv{{\tilde{A}^{-1}_{\vu\vu}}}
\def\BuuCInv{{\tilde{B}^{-1}_{\vu\vu}}}
\def\AvvCInv{{\tilde{A}^{-1}_{\vv\vv}}}
\def\BvvCInv{{\tilde{B}^{-1}_{\vv\vv}}}
\def\AvvInvT{{A^{-\transpose}_{\vv\vv}}}
\def\AuuCInvT{{\tilde{A}^{-\transpose}_{\vu\vu}}}
\def\AvvCInvT{{\tilde{A}^{-\transpose}_{\vv\vv}}}

\def\VXt{{\mathbf{V}^t_{\mX}}}
\def\VXXt{{\mathbf{V}^t_{\mX\mX}}}

\def\dvx{{\delta\vx}}
\def\dvu{{\delta\vu}}
\def\dvv{{\delta\vv}}

\def\vuTraj{{\bar{\vu}}}

\def\Inv{{-1}}

\def\ConvT{{\text{ } \hat{*} \text{ }}}

\def\Qxts{{Q^{t_s}_{\vx}}}
\def\Quxts{{Q^{t_s}_{\vu \vx}}}
\def\Qxuts{{Q^{t_s}_{\vx \vu}}}
\def\Quuts{{Q^{t_s}_{\vu \vu}}}
\def\Qxxts{{Q^{t_s}_{\vx \vx}}}
\def\Qxt{{Q^t_{\vx}}}
\def\Qut{{Q^t_{\vu}}}
\def\Qxxt{{Q^t_{\vx \vx}}}
\def\Quxt{{Q^t_{\vu \vx}}}
\def\Qxut{{Q^t_{\vx \vu}}}
\def\Quut{{Q^t_{\vu \vu}}}

\def\Qu{{Q_{\vu}}}

\def\Quu{{Q_{\vu \vu}}}
\def\QuuInv{{Q^{-1}_{\vu \vu}}}
\def\QuuInvt{{({Q^{t}_{\vu \vu})}^{-1}}}
\def\Qux{{Q_{\vu \vx}}}

\def\Qvt{{Q^t_{\vv}}}
\def\Qyt{{Q^t_{\vx_r}}}
\def\Qyyt{{Q^t_{\vx_r\vx_r}}}
\def\Qyut{{Q^t_{\vx_r\vu}}}
\def\Quyt{{Q^t_{\vu\vx_r}}}
\def\Qyvt{{Q^t_{\vx_r\vv}}}
\def\Qxyt{{Q^t_{\vx\vx_r}}}
\def\Qyxt{{Q^t_{\vx_r\vx}}}
\def\Qvxt{{Q^t_{\vv \vx}}}
\def\Qvyt{{Q^t_{\vv \vx_r}}}
\def\Qxvt{{Q^t_{\vx \vv}}}
\def\Qvvt{{Q^t_{\vv \vv}}}
\def\Quvt{{Q^t_{\vu \vv}}}
\def\Qvut{{Q^t_{\vv \vu}}}
\def\Qv{{Q_{\vv}}}
\def\Qvv{{Q_{\vv \vv}}}
\def\QvvInv{{Q^{-1}_{\vv \vv}}}

\def\Qvx{{Q_{\vv \vx}}}

\def\Quv{{Q_{\vu \vv}}}
\def\Qvu{{Q_{\vv \vu}}}
\def\Qvy{{Q_{\vv \vx_r}}}
\def\QvuT{{Q^\transpose_{\vv \vu}}}
\def\QuvT{{Q^\transpose_{\vu \vv}}}

\def\futsT{{{{f}^{t_s}_{\vu}}^\transpose}}
\def\fxtsT{{{{f}^{t_s}_{\vx}}^\transpose}}
\def\fxts{{{f}^{t_s}_{\vx}}}

\def\fxt{{{f}^t_{\vx}}}
\def\fut{{{f}^t_{\vu}}}
\def\futT{{{{f}^t_{\vu}}^\transpose}}
\def\fxtT{{{{f}^t_{\vx}}^\transpose}}
\def\fxxt{{{f}^t_{\vx \vx}}}
\def\fuut{{{f}^t_{\vu \vu}}}
\def\fuxt{{{f}^t_{\vu \vx}}}

\def\lut{{{\ell}^t_{\vu}}}

\def\luut{{{\ell}^t_{\vu \vu}}}

\def\Vxt{{V^{t+1}_{\vx}}}
\def\Vxxt{{V^{t+1}_{\vx\vx}}}

\def\VXt{{\mathbf{V}^t_{\mX}}}
\def\VXXt{{\mathbf{V}^t_{\mX\mX}}}

\def\transpose{{\mathsf{T}}}

\newcommand{\eq}[1]{{Eq.~(#1)}}

\usepackage{tabularx}
\usepackage{framed}

\usepackage{microtype}
\usepackage{graphicx}
\usepackage{amssymb}
\usepackage{amsthm}
\usepackage{multirow}
\usepackage{siunitx}
\usepackage{algorithmic}
\newtheorem{theorem}{Theorem}

\newtheorem{proposition}[theorem]{Proposition}
\newtheorem{corollary}[theorem]{Corollary}

\usepackage{algorithm}

\let\svthefootnote\thefootnote

\newcommand\numberthis{\addtocounter{equation}{1}\tag{\theequation}}
\usepackage{cancel}
\usepackage{lipsum}

\usepackage[export]{adjustbox}
\usepackage[position=top]{subfig}

\usepackage[T1]{fontenc}
\DeclareCaptionLabelFormat{andtable}{#1~#2  \&  \tablename~\thetable}

\usepackage{hyperref}
\hypersetup{
    colorlinks,
    linkcolor={blue!50!black},
    citecolor={blue!50!black},
    urlcolor={blue!80!black}
}
\usepackage{xcolor}
\usepackage[toc]{appendix}
\usepackage{capt-of}
\usepackage{wrapfig}
\usepackage{color,soul}
\colorlet{color1}{green!50!black}
\colorlet{color2}{orange!95!black}
\colorlet{color3}{red!80!black}
\newcommand{\markgreen}[1]{{\color{color1} #1}}
\newcommand{\markredd}[1]{{\color{color3} #1}}

\usepackage{footnote}
\usepackage{pifont}

\newcommand{\cmark}{\ding{51}}%
\newcommand{\xmark}{\ding{55}}%
\newcolumntype{?}{!{\vrule width 1pt}}

\usepackage{adjustbox}
\usepackage{transparent}
\usepackage{eso-pic}
\usepackage[flushleft]{threeparttable}
\usepackage{mathtools}
\usepackage[prependcaption,textsize=tiny]{todonotes}

\usepackage{titlesec}
\usepackage{amsmath}

\let\oldsqrt\sqrt
\def\sqrt{\mathpalette\DHLhksqrt}
\def\DHLhksqrt#1#2{\setbox0=\hbox{$#1\oldsqrt{#2\,}$}\dimen0=\ht0
\advance\dimen0-0.2\ht0
\setbox2=\hbox{\vrule height\ht0 depth -\dimen0}%
{\box0\lower0.4pt\box2}}

\title{A Differential Game Theoretic Neural \\ Optimizer for Training Residual Networks}

\author{%
  Guan-Horng~Liu,~Tianrong~Chen,~and Evangelos A.~Theodorou\\
  Autonomous Control and Decision Systems Laboratory \\
  Georgia Institute of Technology,Atlanta, GA 30332 \\
  \texttt{\{ghliu,tianrong.chen,evangelos.theodorou\}@gatech.edu} \\
}

\begin{document}

\maketitle

\begin{abstract}
Connections between Deep Neural Networks (DNNs) training and optimal control theory has attracted considerable attention as a principled tool of algorithmic design.
Differential Dynamic Programming (DDP) neural optimizer \cite{liu2020differential} is a recently proposed method along this line.
Despite its empirical success,
the applicability has been limited to feedforward networks
and whether such a trajectory-optimization inspired framework can be extended to modern architectures remains unclear.
In this work, we derive a generalized
DDP optimizer that accepts both residual connections and convolution layers.
The resulting optimal control representation admits a {game theoretic} perspective, in which
training residual networks
can be interpreted as {cooperative trajectory optimization on state-augmented dynamical systems}.
This Game Theoretic DDP (GT-DDP) optimizer
enjoys the same theoretic connection in previous work,
yet generates a much complex update rule that better leverages available information during network propagation.
Evaluation on image classification datasets (e.g. MNIST and CIFAR100) shows an improvement in training convergence and variance reduction
over existing methods.
Our approach highlights the benefit gained from architecture-aware optimization.

\end{abstract}

\section{Introduction}

Attempts from different disciplines to provide a fundamental understanding of deep learning have advanced rapidly in recent years.
Among those, interpretation of DNNs as discrete-time nonlinear dynamical systems, by viewing each layer as a distinct time step,
has received tremendous focus as it enables rich analysis ranging from numerical equations \cite{lu2017beyond},  mean-field theory \cite{schoenholz2016deep}, to physics \cite{shwartz2017opening,greydanus2019hamiltonian,zhong2019symplectic}. 
For instance, 
interpretation of residual networks as a discretization of ordinary differential equations (ODEs) \cite{weinan2017proposal}
provides theoretical reasoning on its optimization landscape \cite{lu2020mean}.
It also inspires new architecture that inherits
numerical stability \cite{sun2018stochastic,chang2018reversible,haber2017stable} and differential limit \cite{chen2018neural,liu2019neural}.

Development of practical optimization methods, however, remains relatively limited. 
This is primarily because classical approach to optimize dynamical systems
relies on the optimal control theory,
which typically considers systems with neither the dimensionality nor parameterization as high as DNNs.
Such a difficulty limits its application,
despite showing promising convergence and robustness in trajectory optimization \cite{tassa2012synthesis},
to mostly theoretical interpretation of DNNs training \cite{han2018mean,liu2019deep}. %
The algorithmic progress has been restricted to either specific network class (e.g. discrete weight \cite{li2018optimal})
or training procedure (e.g. hyper-parameter adaptation \cite{li2017stochastic} or computational acceleration \cite{zhang2019you,gunther2020layer}),
until the recently proposed Differential Dynamic Programming (DDP) optimizer \cite{liu2020differential}.

\begin{wrapfigure}{r}{0.34\textwidth}
  \vskip -0.1in
    \includegraphics[width=0.34\textwidth]{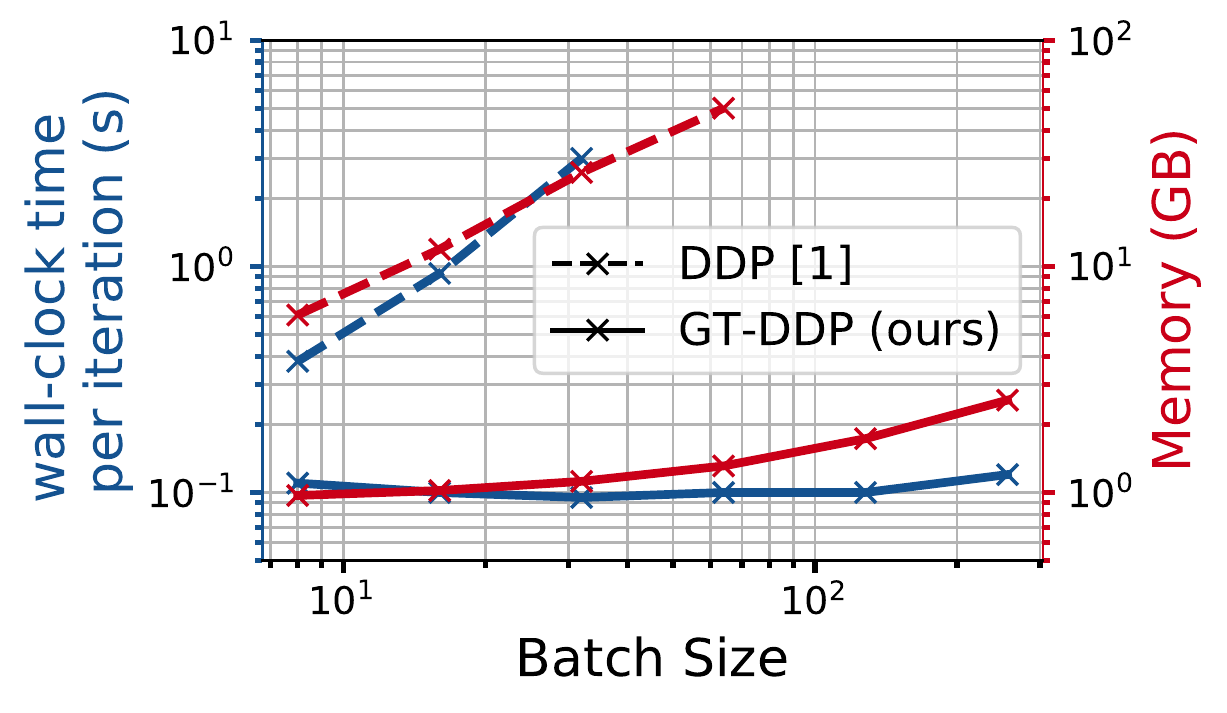}
  \vskip -0.12in
  \caption{
  Comparison on MNIST.
  }
  \label{fig:ddp-compare}
\end{wrapfigure}

DDP is a second-order optimizer built upon
a formal connection between trajectory optimization and training feedforward networks,
and from such it 
suggests existing training algorithms can 
be lifted to embrace 
the dynamic programming principle,
resulting in superior parameter updates with layer-wise feedback policies.
However, DDP is an \textit{architecture-dependent} optimizer, in that 
the feedback policies need to be derived on a per architecture basis.
This raise questions of its flexibility and scalability
to training modern architectures such as residual networks \cite{he2016deep},
since the existing formulation scales exponentially with {the batch size}
(see Fig.~\ref{fig:ddp-compare}).

In this work, we present a game-theoretic extension to the DDP optimizer (GT-DDP) 
which arises naturally from the optimal control representation of residual networks. 
GT-DDP treats each layer as a decision maker in a multi-stage coalition game connected through network propagation.
This leads to much complex feedback policies as information is allowed to exchanged between layers.
Despite the increasing computation, 
we leverage efficient approximations which enable GT-DDP to run 
on a faster wall-clock yet with less memory
(see Fig.~\ref{fig:ddp-compare}).
On the theoretical side,
we extend previous analysis for feedforward networks 
to \textit{arbitrary} architecture (Proposition \ref{prop:bp2ddp2}),
and derive game-theoretic integration for existing second-order optimizers (Theorem \ref{thm:coop} and Corollary \ref{coro:5}).
GT-DDP shows an overall improvement on image classification dataset.

There has been a rising interest in game-theoretic analysis since the landmark Generative Adversarial Network \cite{goodfellow2014generative}.
By framing networks into a two-player competing game,
prevalent efforts have been spent on 
studying its convergence dynamics \cite{kodali2017convergence} and  %
effective optimizers to find stable saddle points \cite{mescheder2017numerics,balduzzi2018mechanics}, 
Notably, our layer-as-player formulation
has appeared in \citet{balduzzi2016grammars}
to study the signal communication implied in the Back-propagation, yet without any practical algorithm being made. 
On the other hand, the cooperative game framework has been used to discover neuron contribution in representation learning and network pruning
\cite{ghorbani2020neuron,stier2018analysing}, 
which are of independent interest for this work.

The paper is organized as follows.
We first review the connection between optimal control and DNNs training in Sec. \ref{sec:prel}.
Extending such framework to residual networks is then given in Sec. \ref{sec:ocp-res},
with GT-DDP demonstrated in Sec. \ref{sec:gt-ddp}. We provide empirical results and discussion in Sec. \ref{sec:experiment} and \ref{sec:discussion}.

\let\thefootnote\relax\footnotetext{
	\textbf{Notation:} $t$ will always be denoted the time step of dynamics, or equivalently the layer's index. 
	Given a time-dependent function $\mathcal{F}_t(\vx_t,\vu_t):\mathbb{X}\times\mathbb{U}\mapsto\mathbb{R}$,
	we will denote and sometimes abbreviate it Jacobian, Hessian, and mixed partial derivative respectively as 
	$\nabla_{\vx_t} \mathcal{F}_t               \equiv \mathcal{F}^t_{\vx}$, 
	$\nabla_{\vx_t}^2 \mathcal{F}_t             \equiv \mathcal{F}^t_{\vx\vx}$, and
	$\nabla_{\vx_t}\nabla_{\vu_t} \mathcal{F}_t \equiv \mathcal{F}^t_{\vx\vu}$.
}
\addtocounter{footnote}{0}\let\thefootnote\svthefootnote

\section{Preliminaries} \label{sec:prel}

\subsection{Optimal Control Formulation of Training DNNs} \label{sec:ocp-dnn}

Classical optimal control problem (OCP) in discrete time considers the following programming:
\begin{equation}
\min_{\vuTraj} J(\bar{\vu}; \vx_0) :=
\left[
    \phi(\vx_{T}) + \sum_{t=0}^{T-1}\ell_t(\vx_{t}, \vu_{t})
\right]
\quad \text{s.t. } \vx_{t+1} = f_t(\vx_{t}, \vu_{t}) \comma
\label{eq:ocp}
\end{equation}
where
$\vx_t \in \mathbb{R}^{n_t} $ and $\vu_t \in \mathbb{R}^{m_t}$
represent the state and control
at each time step $t \in \{0,\cdots,T\}$.
$f_t$, $\ell_t$ and $\phi$ respectively denote the dynamics, intermediate cost and terminal cost.
The control trajectory is denoted as $\vuTraj \triangleq \{\vu_t\}_{t=0}^{T-1}$.
\eq{\ref{eq:ocp}} can be interpreted as the training objective of DNNs by treating
$\vx_t$ and $\vu_t$ as the vectorized activation map
($\vx_0$ and $\vx_T$ being input image and prediction vector) 
and weight at each layer $t$.
$f_t$ stands as the compositional module
propagating the activation vector,
e.g. an affine transformation followed by an element-wise activation in a feedforward network.
$\ell_t$ and $\phi$ denote the per-layer regularization (e.g. weight decay) and terminal loss (e.g. cross-entropy).

Following these notations, the gradient descent (GD) update at iteration $k$ can be written as
$\vuTraj^{(k+1)}= \vuTraj^{(k)}+\delta \vuTraj^* = \vuTraj^{(k)}-\eta \nabla_{\vuTraj} J$,
where $\eta$ is the learning rate.
We can further break down the update for the full network to each layer, i.e.
$\delta \vuTraj \triangleq \{\delta \vu_t\}_{t=0}^{T-1}$,
computed backward by
\begin{align}
\dvu_t^*
&= \argmin_{\dvu_t \in \mathbb{R}^{m_t}}\{
	J_t +  \nabla_{\vu_t} J_t^\transpose \dvu_t + \textstyle \frac{1}{2} \dvu_t^\transpose (\textstyle \frac{1}{\eta}\mI_t) \dvu_t
\} \comma \label{eq:du-star} \\
\text{where } J_t(\vx_t,\vu_t) &\triangleq \ell_t(\vu_t) + J_{t+1}(f_t(\vx_t,\vu_t),\vu_{t+1})
\comma \quad J_T(\vx_T)\triangleq\phi(\vx_T)
\label{eq:Jt}
\end{align}
is the per-stage objective\footnote{
	We drop $\vx_t$ in all $\ell_t(\cdot)$ hereafter
	as the layer-wise regularization typically involves network weight alone.
} at layer $t$.
It can be readily verified that $\nabla_{\vx_t}J_t$ gives the exact Back-propagation dynamics.
\eq{\ref{eq:du-star}} follows the standard optimization interpretation in which GD minimizes the second-order Taylor expansion of $J_t$
with its Hessian $\nabla_{\vu_t}^2 J_t$ replaced by $\frac{1}{\eta}\mI_t$, i.e. spherical curvature.
In a similar vein,
adaptive first-order methods, such as RMSprop and Adam,
approximate $\nabla_{\vu_t}^2 J_t$ by diagonal matrices
with the leading entries
adapting to the second-moment statistics in each coordinate.
Second-order optimizers like KFAC \cite{martens2015optimizing,grosse2016kronecker} and EKFAC \cite{george2018fast}
compute much complex non-diagonal curvature matrices with Gauss-Newton approximation,
i.e. $\nabla_{\vu_t}^2 J_t \approx J^t_\vu {J^t_\vu}^\transpose$.

\begin{figure}
\begin{minipage}[t]{0.49\textwidth}
\vskip -0.25in
\begin{algorithm}[H]
\small
     \caption{\small DDP Neural Optimizer (at iteration $k$)}
     \label{alg:ddp-no}
  \begin{algorithmic}[1]
   \STATE {\bfseries Input:} forward pass $\{ {\vx}_t \}_{t=0}^{T}$ with weights $\bar{\vu}^{(k)}$
   \STATE Set $V_\vx^{T} = \nabla_\vx \phi$ and $V_{\vx \vx}^{T} = \nabla^2_{\vx} \phi$
   \FOR{$t=T-1$ {\bfseries to} $0$}
   \STATE Compute derivatives of $Q_t$ with $V_{\vx}^{t+1}$, $V_{\vx \vx}^{t+1}$
   \STATE Compute $\kt$, $\Kt$, $V_{\vx}^t$ and $V_{\vx \vx}^t$
   \ENDFOR
   \STATE Set $\hat{\vx}_0 = {\vx}_0$
   \FOR{$t=0$ {\bfseries to} $T-1$}
   \STATE $\hat{\vu}_t = {\vu}_t^{(k)} + \kt + \Kt \dvx_t, \text{ }\text{ }\text{ }(\dvx_t=\hat{\vx}_t - {\vx}_t)$
   \STATE $\hat{\vx}_{t+1} = f_t(\hat{\vx}_t, \hat{\vu}_t)$
   \ENDFOR
   \STATE $\vuTraj^{(k+1)} \leftarrow \{ \hat{\vu}_t \}_{t=0}^{T-1}$
  \end{algorithmic}
\end{algorithm}
\end{minipage}
\hfill
\begin{minipage}[t]{0.49\textwidth}
\vskip -0.07in
\centering
\includegraphics[width=\textwidth]{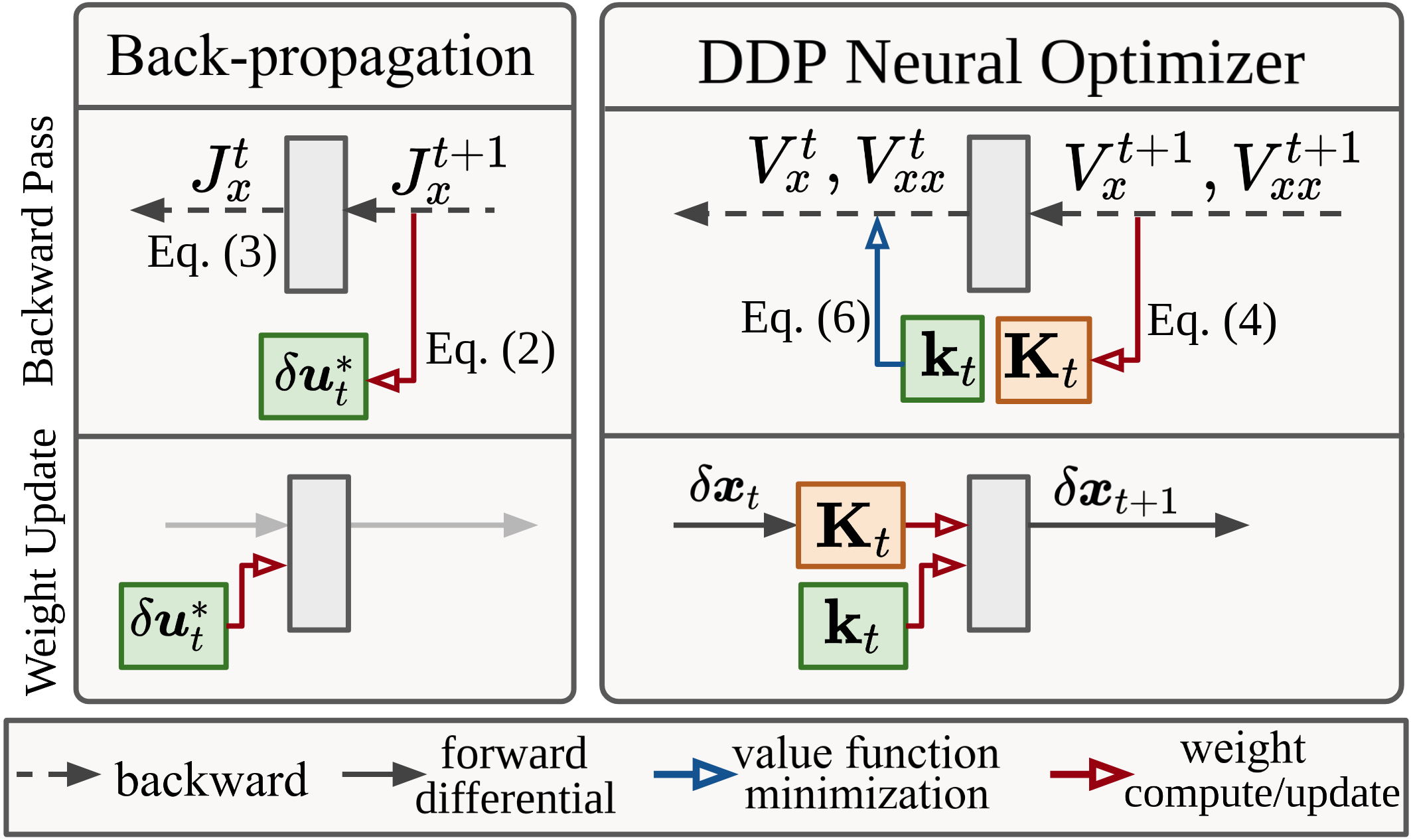}
\caption{Comparison of computational graphs in feedforward networks.
	}
\label{fig:ddp-no}
\end{minipage}
\vskip -0.17in
\end{figure}

\subsection{Differential Dynamic Programming Neural Optimizer} \label{sec:ddp}
Differential Dynamic Programming (DDP) is a second-order trajectory optimization algorithm that solves the same programming in \eq{\ref{eq:ocp}}.
Instead of searching updates from $\mathbb{R}^{m_t}$,
at each decision stage DDP aims at finding a locally-optimal \textit{feedback} policy,
i.e. $\dvu_t(\dvx_t) \in \Gamma_{\dvx_t}$, where
$\Gamma_{\dvx_t} = \{\mathbf{b}_t + \mathbf{A}_t \dvx_t: \mathbf{b}_t \in \mathbb{R}^{m_t}, \mathbf{A}_t \in \mathbb{R}^{{m_t}\times {n_t}} \}$
denotes all possible {affine} mappings from the state differential $\dvx_t$.
The resulting per-stage updates can also be computed backward:
\begin{align}
\dvu_t^* (\dvx_t)
= \argmin_{\dvu_t \in \Gamma_{\dvx_t}} &\{
	Q_t +
	\frac{1}{2}
	\left[\begin{array}{c} {\mathbf{1}} \\ {\dvx_t} \\ {\dvu_t} \end{array}\right]^{\transpose}
	\left[\begin{array}{ccc}
		{\mathbf{0}}    & {\Qxt^\transpose} & {\Qut^\transpose} \\
		{\Qxt} & {\Qxxt} & {\Qxut} \\
		{\Qut} & {\Quxt} & {\Quut}
		\end{array}\right]
	\left[\begin{array}{c} {\mathbf{1}} \\ {\dvx_t} \\ {\dvu_t} \end{array}\right]
	\} \comma \label{eq:du-star-ddp}
\\
\text{where }
	V_t(\vx_t) \triangleq \min_{\vu_t \in \Gamma_{\vx_t}}&
		\underbrace{
		\ell_t(\vu_t) + V_{t+1}(f_t(\vx_t,\vu_t))
		}_{Q_t(\vx_t,\vu_t)\equiv Q_t}
	\comma \quad V_T(\vx_T)\triangleq\phi(\vx_T)
\label{eq:Qt}
\end{align}
is the \textit{value} function that summarizes the objective value when all the afterward stages, i.e. $Q_{s \ge t}$,
are minimized.
Hereafter we will denote the quadratic expansion in \eq{\ref{eq:du-star-ddp}} as $\delta Q_t(\dvx_t,\dvu_t)$.
$Q_t$ will be referred to the \textit{Bellman objective}, as \eq{\ref{eq:Qt}} is well-known as the \textit{Bellman equation}
\cite{bellman1954theory}.

The analytic solution to \eq{\ref{eq:du-star-ddp}} is given by
$\dvu^{*}_t(\dvx_t) = \kt + \Kt \dvx_t$,
where
$\kt \triangleq -(\Quut)^\Inv\Qut$ and $\Kt \triangleq -(\Quut)^\Inv\Quxt$
are the locally optimal open and feedback gains.
From the chain rule, evaluating the derivatives of $Q_t$ in \eq{\ref{eq:du-star-ddp}} requires one to compute
$V^{t+1}_{\vx}$ and $V^{t+1}_{\vx \vx}$.
These quantities can be obtained by simply substituting $\dvu^{*}_t(\dvx_t)$ to \eq{\ref{eq:Qt}} at each stage:
\begin{align} \label{eq:value-recursive}
\begin{split}
V_{\vx}^t     &= \nabla_{\vx_t}   \{Q_t+\delta Q_t(\dvx_t,\dvu^{*}_t(\dvx_t))\} = \Qxt  + \Qxut \kt \comma \\
V_{\vx \vx}^t &= \nabla_{\vx_t}^2 \{Q_t+\delta Q_t(\dvx_t,\dvu^{*}_t(\dvx_t))\} = \Qxxt + \Qxut \Kt \period
\end{split}
\end{align}

It is obvious that Eq.~(\ref{eq:du-star-ddp}, \ref{eq:Qt})
resemble Eq.~(\ref{eq:du-star}, \ref{eq:Jt})
in several ways.
Both classes of optimizer perform quadratic approximation of the stage-wise objective,
except DDP also expands the objective wrt $\dvx_t$, which requires computing the mixed partial derivatives $\Quxt$.
The theoretical connection between these two approaches for feedforward networks
has been made formally in \citet{liu2020differential}.
\begin{proposition}[\cite{liu2020differential}] \label{prop:bp2ddp}
When $Q_{\vu \vx}^t=\mathbf{0}$ at all stages,
the first-order derivative of the value function collapses to the Back-propagation gradient in feedforward networks,
i.e. $V^t_{\vx} = J^t_{\vx}$.
In this case, DDP computes the same update in stage-wise Newton\footnote{
	Stage-wise Newton preconditions the gradient by the block-wise inverse Hessian at each layer.
}:
$\dvu^{*}_t(\dvx_t) = - (J^t_{\vu\vu})^{\Inv}J^t_{\vu}$.
If we further assume ${Q_{\vu \vu}^{t}} = \frac{1}{\eta}\mI_{t}$,
then DDP degenerates to the Back-propagation with gradient descent.
\end{proposition}
Proposition \ref{prop:bp2ddp} suggests that
by setting $\Quxt=\mathbf{0}$ and choosing a proper $\Quut$, we can recover existing optimizers from DDP.
Meanwhile, existing methods can be extended to accept DDP framework by computing $\Quxt$.
The resulting layer-wise feedback policies generate weight update with additional forward pass (lines $7$-$11$ in Alg.~\ref{alg:ddp-no}),
in which the state differential is computed.
We summarize the backward pass and weight update procedure of the DDP optimizer in Alg.~\ref{alg:ddp-no} and Fig.~\ref{fig:ddp-no}.

\section{Optimal Control Representation for Residual Networks} \label{sec:ocp-res}

In this section, we extend the Bellman optimization framework to networks
consist of residual paths.
Despite that in the Back-propagation this simply involves merging additional gradient flow from the shortcut,
its optimal control representation is much complex when
second-order information and Bellman minimization are involved.
We leave the complete derivation in the Appendix \ref{app:sec3-dev}.

\newsavebox{\mycases}%

\subsection{Residual Connection as State-Augmented Dynamics} \label{sec:31}

\begin{figure}[h]%
  \vskip -0.15in
  \hspace{-0.12in}
  \subfloat{
  	\includegraphics[width=0.49\linewidth]{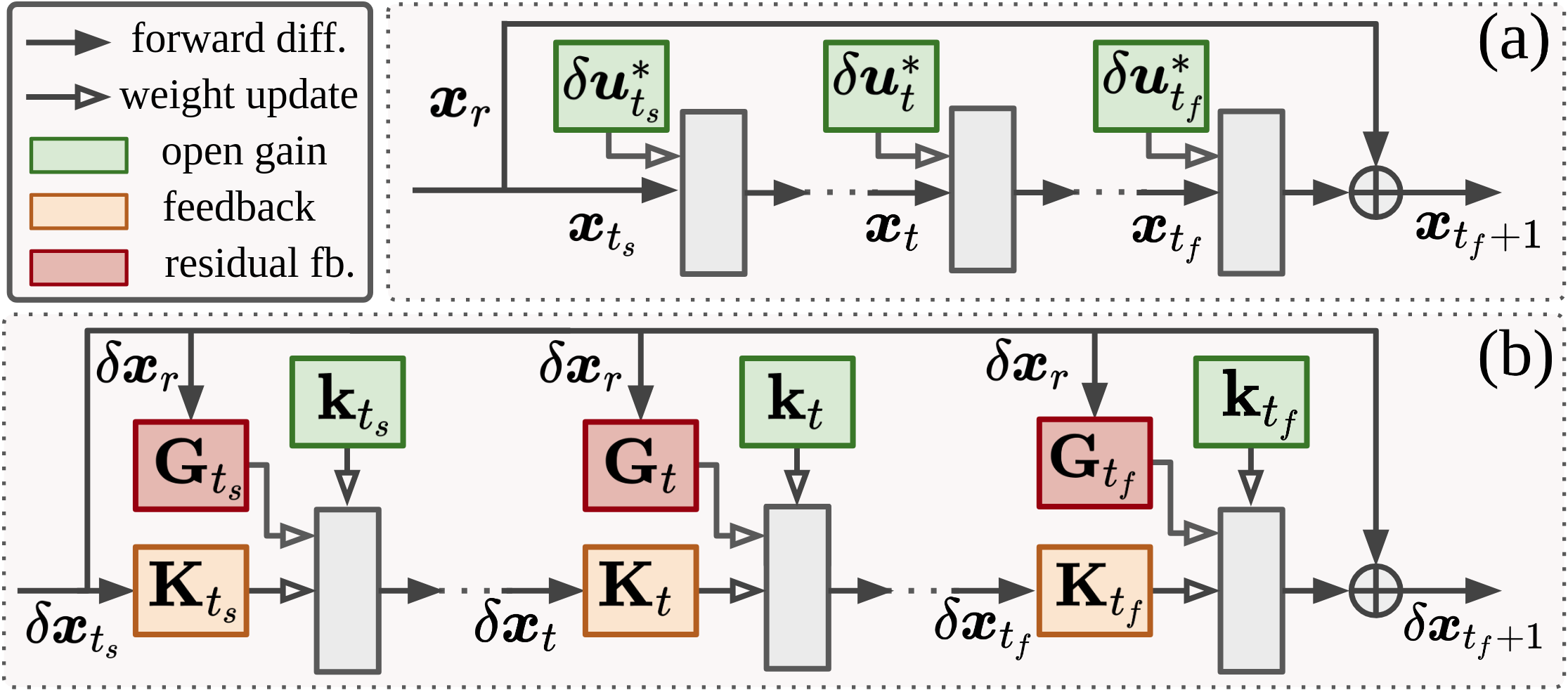}
  	\label{fig:weight-update-a}%
  }
  \subfloat{
  	\includegraphics[width=0.52\linewidth]{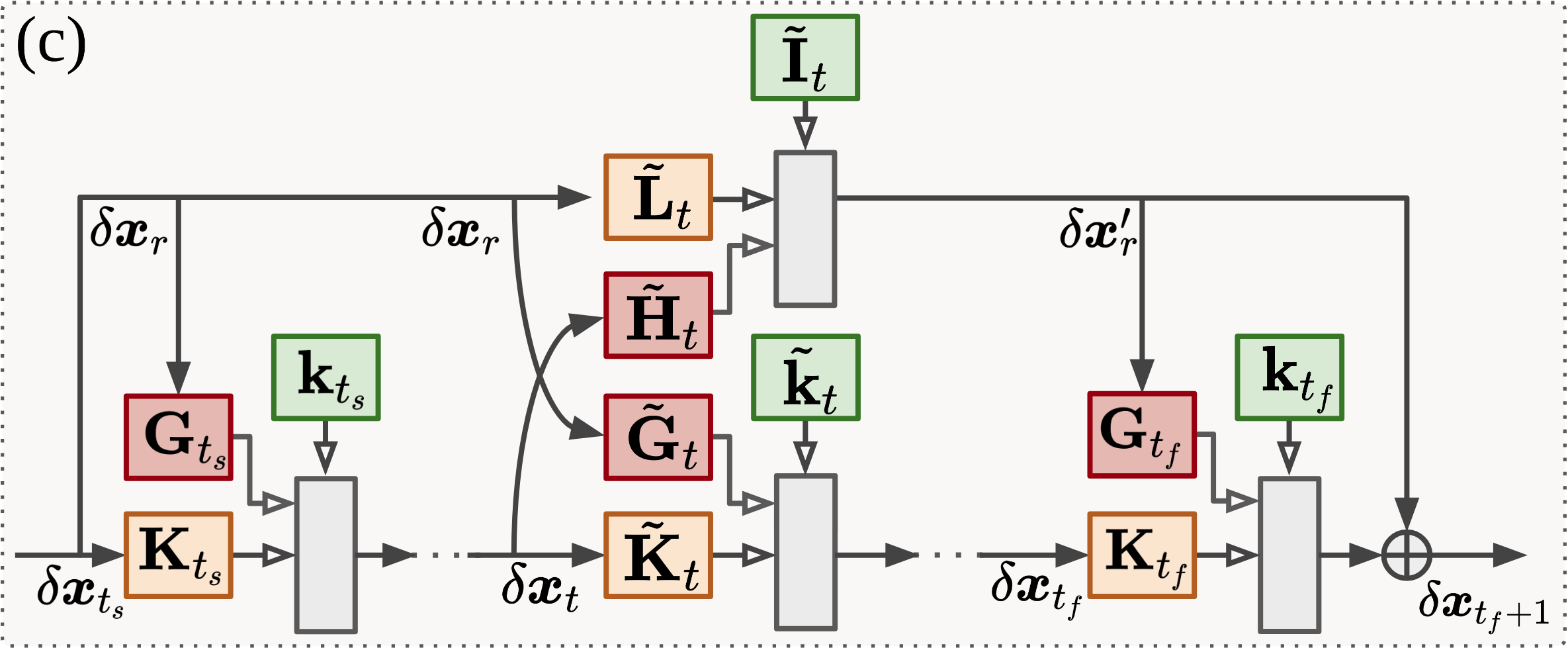}
  	\label{fig:weight-update-b}%
  }
  \subfloat{
  	\textcolor{white}{\rule{1pt}{1pt}}
  	\label{fig:weight-update-c}%
  }
  \caption{
  	Terminology and weight update graph for (a) standard Back-propagation and (b)(c) GT-DDP optimizer with identity and arbitrary
  	shortcut mapping.
  }
  \hspace{-0.2in}
  \label{fig:weight-update}%
  \vskip -0.2in
\end{figure}

Consider the residual network in Fig.~\ref{fig:weight-update-a}.
Let us denote $\vx_r$ as the residual state shortcutting from the layer $t_s$ to $t_f$,
{so that the output is merged by $\vx_{t_f+1} = \vx_r + f_{t_f}(\vx_{t_f},\vu_{t_f})$}.
The Bellman equation along the residual path is given by
\begin{align}
V_{t_s}(\vx_{t_s}) = \textstyle \min_{\vu_{t\in[t_s,t_f] }}
	\ell_{t_s}(\vu_{t_s}) + \cdots + \ell_{t_f}(\vu_{t_f})
	+ V_{t_f+1
	}(\vxr+
		(f_{t_f} \circ \cdots \circ f_{t_s})(\vx_{t_s})
	)
\comma
\label{eq:value123}
\end{align}
which can be decomposed into the following minimization and solve recursively from $t_f$:
\begin{subequations} \label{eq:values}
\begin{align}
  \sbox{\mycases}{$V_{t}(\vxr,\vx_{t}) = \textstyle \min_{\vu_{t}} Q_{t}(\vxr,\vx_{t},\vu_{t}) :=
  \left\{\begin{array}{@{}c@{}}\vphantom{ \ell_{t}(\vu_{t}) + V_{t+1}(\vxr + f_{t}(\vx_{t},\vu_{t})) \comma \quad t = t_f}\\\vphantom{\ell_{t}(\vu_{t}) + V_{t+1}(\vxr, f_{t}(\vx_{t},\vu_{t})) \comma \quad t \in (t_s,t_f)}\end{array}\right.\kern-\nulldelimiterspace$}
  \raisebox{-.5\ht\mycases}[0pt][0pt]{\usebox{\mycases}}
  	 \ell_{t}(\vu_{t}) + V_{t+1}(\vxr + f_{t}(\vx_{t},\vu_{t})) \comma \quad t = t_f \quad       \label{eq:value1} \\
     \ell_{t}(\vu_{t}) + V_{t+1}(\vxr, f_{t}(\vx_{t},\vu_{t})) \comma \quad t \in (t_s,t_f)	\label{eq:value2} \\
V_{t_s}(\vx_{t_s})      = \textstyle \min_{\vu_{t_s}} Q_{t_s}(\vx_{t_s},\vu_{t_s})
	:= \ell_{t_s}(\vu_{t_s}) + V_{t_s+1}(\vx_{t_s},f_{t_s}(\vx_{t_s},\vu_{t_s})) \qquad \qquad \label{eq:value3}
\end{align}
\end{subequations}
\eq{\ref{eq:values}} suggests
the value functions of layers parallel to the shortcut
depend not only on its own state $\vx_t$ but also the residual $\vx_r$.
This is better explained from the game theoretic viewpoint.
As $\vx_r$ affects the payoff obtained during $t \in [t_s,t_f]$ through {the addition at $t_f+1$},
it shall contribute to decisions made at these stages.
Notice that we can rewrite the propagation rule as state-augmented dynamics $\hat{f}_t(\vxr,\vx_t,\vu_t)$.
Dynamics of such forms resemble \textit{time-delayed} systems \cite{fan2016differential},
$f(\vx_{t-i},\cdots,\vx_{t},\vu_{t})$.
Instead of a constant moving window, here we consider a fixed time stamp anchored at $t_s$.

The new DDP update can be solved similar to \eq{\ref{eq:du-star-ddp}},
except the Bellman objective should be expended additionally wrt to  $\dvx_r$.
The optimal feedback law thus depends on the differential of both states:
\begin{align}
	\dvu_t^* (\dvx_t,\dvx_r) = \kt + \Kt \dvx_t + \Gt \dvx_r \comma \quad
	\text{where } \Gt \triangleq -(\Quut)^\Inv {f^t_{\vu}}^\transpose V^{t+1}_{\vx\vxr}
	\label{eq:du-res}
\end{align}
is the optimal residual feedback gain.
$\kt$ and $\Kt$ are the same open and feedback gains computed in the absence of shortcut.
Thus, the new update rule has an additional feedback from the channel of residual state (cf. Fig.~\ref{fig:weight-update-b}).
The term
$V^{t+1}_{\vx\vxr}$
denotes the mixed partial derivatives of $V_{t+1}(\vxr,\vx_{t+1})$,
quantifying how these two states should be correlated mathematically.
It can be computed, together with the residual value Hessian $V^{t+1}_{\vxr\vxr}$,
through backward recursions similar to \eq{\ref{eq:value-recursive}},
\begin{align}
	V^{t}_{\vx\vxr} = {f^t_{\vx}}^\transpose V^{t+1}_{\vx\vxr} - \KtT \Quut \Gt
	\comma \quad
	V^{t}_{\vxr\vxr} = V^{t+1}_{\vxr\vxr} - \GtT \Quut \Gt
	\comma
	\label{eq:vxx_res}
\end{align}
with the terminal conditions given by
$V^{t_f+1}_{\vx\vxr} = V^{t_f+1}_{\vxr\vxr} = V^{t_f+1}_{\vx\vx}$.

It is natural to ask how the optimal control representation differs between residual and feedforward networks.
This is summarized in the following proposition.
\begin{proposition} \label{prop:res}
When networks contain shortcut from $t_s$ to $t_f$,
the derivatives of the value function at stage $t_s$,
denoted $\tilde{V}^{t_s}_{\vx}$ and $\tilde{V}^{t_s}_{\vx\vx}$,
relate to the ones in feedforward networks, denoted ${V}^{t_s}_{\vx}$ and ${V}^{t_s}_{\vx\vx}$, by
\begin{align}
	\tilde{V}^{t_s}_{\vx}    &= V^{t_s}_{\vx}    + V^{t_f+1}_{\vx} - \textstyle \sum_{t \in [t_s, t_f]} \GtT\Quut\kt
	\comma \label{eq:vx-res} \\
	\tilde{V}^{t_s}_{\vx\vx} &= V^{t_s}_{\vx\vx } + V^{t_f+1}_{\vx\vx} - \textstyle \sum_{t \in [t_s, t_f]} \GtT\Quut\Gt
							  + V^{t_s}_{\vx\vxr} + {V^{t_s\transpose}_{\vx\vxr}} \label{eq:vxx-res2}
\end{align}
\end{proposition}
There are several interesting implications from Proposition \ref{prop:res}.
First, recall that in the Back-propagation,
the gradient at $t_s$ is obtained by simply merging the one from the shortcut,
i.e. $\tilde{J}^{t_s}_{\vx} = J^{t_s}_{\vx} + J^{t_f+1}_{\vx}$.
In the Bellman framework, $\tilde{V}^{t_s}_{\vx}$ is modified in a similar manner, yet with an additional summation
coming from the Bellman minimization along the shortcut.
Interpretation for $\tilde{V}^{t_s}_{\vx\vx}$
follows the same road map, except the mixed partial derivative $V^{t_s}_{\vx\vxr}$ also contributes to {the Hessian of the value function at $t_s$}. %
We highlight these traits which distinguish our work
from both standard Back-propagation and previous work \cite{liu2020differential}.

\setul{0.9ex}{0.2ex}
\setulcolor{color1}

\subsection{Cooperative Trajectory Optimization with Non-identity Shortcut Mapping} \label{sec:32}
In some cases,
the dimension of feature map between $t_s$ and $t_f$ may be mismatched; %
thus the residual path will contain a non-identity shortcut mapping \cite{he2016identity}.
For CNNs this is typically achieved by down-sampling $\vx_{r}$ with an 1$\times$1 convolution.
Hereafter we will denote this non-identity mapping as $\vx_r^\prime=h_{t}(\vx_r,\vv_t)$,
where $\vv_t$ is the vectorized weight.
The new Bellman equation, consider for instance when
we add the mapping to the middle of residual path, i.e. $t\in(t_s,t_f)$ in \eq{\ref{eq:value2}},
becomes
\begin{align}
V_t(\vx_r,\vx_t) = \min_{\vu_t,\vv_t}
	\underbrace{
		\ell(\vu_t)+\ell(\vv_t) + V_{t+1}(h_t(\vx_r,\vv_t),f_t(\vx_t,\vu_t))
	}_{\triangleq Q_t(\vx_r,\vx_t,\vu_t,\vv_t)}
\period
\label{eq:coop-traj}
\end{align}
Minimizing $Q_t(\vx_r,\vx_t,\vu_t,\vv_t)$ simultaneously wrt $\vu_t$ and $\vv_t$ resembles the formulation in
a complete Cooperative Game (CG) \cite{yeung2006cooperative}.
In its common setup,
two players observe the same state and decide their policies to maximize a {\textit{cooperative} payoff.}
The game is \textit{complete} in that all information
is known and shared in prior; thus can be leveraged to make better decisions.
Application of DDP to solving CG has been studied previously in robotics for robust trajectory optimization \cite{pan2015robust}.

Before solving \eq{\ref{eq:coop-traj}},
it will be useful to first revisit cases when each policy can be solved independently,
i.e. when $Q_t(\vx_r,\vx_t,\vu_t,\vv_t)=Q_t(\vx_t,\vu_t)+Q_t(\vx_r,\vv_t)$.
In this case, we know $\kt,\Kt$ is the solution to $\argmin_{\vu_t} Q_t(\vx_t,\vu_t)$.
Let us further denote $\It+\Lt\dvx_r=\argmin_{\vv_t} Q_t(\vx_r,\vv_t)$,
where
$\It \triangleq -(\Qvvt)^\Inv\Qvt$ and $\Lt \triangleq -(\Qvvt)^\Inv\Qvyt$.
Now, solving \eq{\ref{eq:coop-traj}} by quadratically expanding $Q_t(\vx_r,\vx_t,\vu_t,\vv_t)$ wrt all variables
will arrive at the following form\footnote{
	We omit the superscript $t$ of $\QuuInv$,$\QvvInv$,$\QuuCInv$,$\QvvCInv$ sometimes for notational simplicity but
  stress that $Q$ is always time (i.e. layer) dependent in this work.
}:
\begin{align*} %
	  \dvu_t^*(\dvx_t,\dvx_r)
	= & \textcolor{color1}{\kut} + \textcolor{color2}{\Kut} \dvx_t + \textcolor{color3}{\Gut} \dvx_r \numberthis \label{eq:du-coop}\\
	= & -\QuuCInv
	  \Big(
	  	\text{\ul{$\Qut+\Quvt\It$}} + \setulcolor{color2}
	  	\text{\ul{$(\Quxt-\Quvt\QvvInv\QBt)$}} \dvx_t + \setulcolor{color3}
	  	\text{\ul{$(\QAt+\Quvt\Lt)$}} \dvx_r
	   \Big) \comma \\
	  \dvv_t^*(\dvx_t,\dvx_r)
	= & \textcolor{color1}{\Ivt} + \textcolor{color2}{\Lvt} \dvx_r + \textcolor{color3}{\Hvt} \dvx_t \numberthis \label{eq:dv-coop}\\
	= & -\QvvCInv
	  \Big(
	  	\text{\ul{$\Qvt+\Qvut\kt$}} + \setulcolor{color2}
	  	\text{\ul{$(\Qvyt-\Qvut\QuuInv\QAt)$}} \dvx_r + \setulcolor{color3}
	  	\text{\ul{$(\QBt+\Qvut\Kt)$}} \dvx_t
	   \Big),
\end{align*}
where
$\QuutC \triangleq \Quut-\Quvt \QvvInv \Qvut$ and
$\QvvtC \triangleq \Qvvt-\Qvut \QuuInv \Quvt$
result from the block-matrices inversion with the Schur complement.
The update rules provided in Eq.~(\ref{eq:du-coop}, \ref{eq:dv-coop})
are much complex and do not admit forms of superposition
as in \eq{\ref{eq:du-res}}.
To make some intuitions,
compare for instance the open gain $\kt \triangleq -\QuuInv\Qut$ with its cooperative variant $\kut \triangleq -\QuuCInv(\Qut+\Quvt\It)$.
The latter adjusts the policy by knowing the companion's update rule $\It$,
and information between two players' actions communicates through $\Quv$ and $\Qvu$.
Similar interpretation can be drawn for the feedback gains $\Kt$ and $\Kut$,
as $\Qvu$ allows information to flow from $\Qux$ through $\Qvx$, and etc.

Fig.~\ref{fig:weight-update-c} illustrates how these feedback policies generate the weight update.
$\kt$ and $\Kt$ are applied in the same manner as in feedforward networks (c.f. Fig.~\ref{fig:ddp-no}).
Layers parallel to the skip connection receive additional residual feedback from $\Gt$.
At the decision stage when the non-identity shortcut mapping is involved,
policies will be modified to their cooperative form, i.e. $\kut, \Kut, \Gut, \Ivt, \Lvt, \Hvt$.
Notice that the residual policies $\mathbf{G}_{s\le t}$ and $\mathbf{G}_{s>t}$
now take \textit{different} state differential ($\delta \vx_r$ and $\delta \vx^\prime_r$ resp.).
This implies the GT-DDP solution to residual networks is not unique,
as placing $h_t(\vx_r,\vv_t)$ at different location along the shortcut will result in different value of weight update.
Despite seemly unintuitive,
from the game theoretic perspective it implies
one would prefer $\delta \vx^\prime_r$ to $\delta \vx_r$ whenever the former is available,
since
states closer to the decision stage reveal more information.

\section{Game Theoretic DDP Neural Optimizer} \label{sec:gt-ddp}

\begin{savenotes}
\begin{table}[t]
  \vskip -0.1in
  \captionsetup{justification=centering}
  \caption{Relation between existing first (e.g. RMSprop) and}
  \label{table:1}
  \centering
      \begin{small}
      \begin{tabular}{r?cc|c}
      \multicolumn{4}{c}{\normalsize second-order (e.g. EKFAC) algorithms under GT-DDP framework\footnote{
      $\odot$ denotes element-wise multiplication. $\vh$ is the pre-activation vector defined in Sec. \ref{sec:gt-ddp}.}
      }\\[3pt]
      \toprule
      & \multirow{2}{*}{$\Quu$,$\Qvv$} & \multirow{2}{*}{$\Quv$,$\Qvu$} & nonzero $\Qux$, \\
      &                                &                                & $\Qvy$,$\QA$, $\QB$ \\
      \midrule
      RMSprop        & $\frac{1}{\eta}\diag({J}_\vu \odot {J}_\vu +\epsilon)$ & $\mathbf{0}$ & \xmark \\
      GT-DDP-RMSprop & $\frac{1}{\eta}\diag({Q}_\vu \odot {Q}_\vu +\epsilon)$ & $\mathbf{0}$ & \cmark \\
      EKFAC        & {$\E[\vx\vx^\transpose] \otimes \E[{J_\vh}J_\vh^\transpose]$} & $\mathbf{0}$ & \xmark \\
      GT-DDP-EKFAC & {$\E[\vx\vx^\transpose] \otimes \E[{V_\vh}V_\vh^\transpose]$} & Theorem \ref{thm:coop} & \cmark \\
      \bottomrule
      \end{tabular}
      \end{small}
      \vskip -0.15in
\end{table}
\end{savenotes}

In this section we discuss efficient computation of
the update rules proposed in the previous section
to training residual networks.
As the algorithm generalizes the DDP framework \cite{liu2020differential} to new architectures under {game-theoretic perspective},
we name it the Game Theoretic DDP neural optimizer (GT-DDP).
Detailed derivation and proof in this section are left in the Appendix \ref{app:sec4-dev}.

\subsection{Curvature Approximation}

Computation of the GT-DDP solution involves extensive evaluation of
the {derivatives of $Q_t$} wrt different variables.
Since $f_t$ is highly over-parametrized in each network layer,
second-order derivatives wrt the weight parameter, e.g. $\Quut$ $\Qvvt$,
are particularly expansive to compute,
let alone their inversions.
Thus, approximation must be made for these matrices.

Following the curvature interpretation in Sec \ref{sec:ocp-dnn},
one can simply substitute these expansive Hessians
with the ones considered in existing methods. %
For instance, replacing $\Quut$ with an identity (or diagonal) matrix resembles the (adaptive) first-order update rule.
Note that this first-order approximation implicitly implies
both $\Quvt$ and $\Qvut$
to vanish,
since by construction first-order methods
omit the covariances among different weight coordinate.

As for second-order approximation,
in this work we consider the popular Kronecker factorization used in EKFAC \cite{george2018fast}.
Let $f_t \equiv \sigma(\mW_t \vx_t $+$ \vb_t)$ be the generic dynamics where $\sigma$ is the activation function,
and denote $\vh_t \equiv \mW_t \vx_t+\vb_t$ as the pre-activation vector.
EKFAC factorizes
$Q_{\vu\vu}^t \approx \E[\vx_t\vx_t^\transpose] \otimes \E[\vg_t\vg_t^\transpose]$,
where
$\otimes$ is the Kronecker operator and
{$\vg_t :=J^t_{\vh}$} is {the first-order derivative of the per-stage objective}
wrt the pre-activation vector\footnote{
  \label{ft:h}
  For GT-DDP-EKFAC, we have $\vg_t := V^t_{\vh}$.
  We left further introduction and derivation in Appendix \ref{app:pre-kron}.
}.
The expectation is taken wrt the batch sample.
Factorizing GT-DDP with Kronecker operation requires one to derive
the Kronecker representations for the cooperative matrices appeared in CG,
which are given below.
\begin{theorem}[Kronecker factorization in Cooperative Game] \label{thm:coop}
Suppose $\Quu$ and $\Qvv$ are factorized respectively by
$Q_{\vu\vu} \approx \Auu \otimes \Buu$ and
$Q_{\vv\vv} \approx \Avv \otimes \Bvv$, where
\begin{align*}
  \Auu \triangleq \Axx \comma \quad \Buu \triangleq \Bxx \comma \quad
  \Avv \triangleq \Ayy \comma \quad \Bvv \triangleq \Byy
\end{align*}
are the Kronecker block matrices for layers $f(\vx_\vu,\vu)$ and $h(\vx_\vv,\vv)$.
Further, let
$\Auv \triangleq \Axy$ and
$\Buv \triangleq \Bxy$,
then the unique Kronecker factorizations for the matrices in CG are given by
  \begin{align}
  \QuuCInv &\approx \AuuCInv \otimes \BuuCInv = (\Auu-\Auv\AvvInv\AuvT)^\Inv \otimes (\Buu-\Buv\BvvInv\BuvT)^\Inv \label{eq:thm1} \\
  \QvvCInv &\approx \AvvCInv \otimes \BvvCInv = (\Avv-\AuvT\AuuInv\Auv)^\Inv \otimes (\Bvv-\BuvT\BuuInv\Buv)^\Inv \label{eq:thm2} \comma
  \end{align}
and $\Quv = \QvuT \approx - \Auv \otimes \Buv$.
The CG update, take $\kut$ for example, can be computed by
  \begin{align}
  \kut
  =-\vectorize(\BuuCInv(\Qu+\Buv\BvvInv \Qv \AvvInvT \AuvT  )\AuuCInvT)
  \period
  \end{align}
\end{theorem}
Hereafter
we will refer these approximations respectively to \textit{GT-DDP-RMSprop}, \textit{GT-DDP-EKFAC}, and etc.
The algorithmic relation between existing methods and their DDP integration is summarized in Table~\ref{table:1},
with the theoretical connection given by the following proposition.
\begin{proposition} \label{prop:bp2ddp2}
The update rules derived from stage-wise minimization of the Bellman equation degenerate to
the method it uses to approximate the weight Hessian, i.e. $\Quu$ $\Qvv$,
when the Bellman objective $Q_t$ at all stages satisfies
{\normalfont(i)} all mixed partial derivatives between parameter and activation, e.g. $\Qux$,$\QA$, vanish,
and
{\normalfont(ii)} parameters between distinct layers are uncorrelated.
\end{proposition}
Note that Proposition \ref{prop:bp2ddp2} extends Proposition \ref{prop:bp2ddp} to \textit{arbitrary} architectures beyond feedforward and residual networks,
so long as its layer-wise Bellman objective is properly defined.

\subsection{Practical Implementation}

\begin{wrapfigure}{r}{0.25\textwidth}
  \vskip -0.31in
    \includegraphics[width=0.25\textwidth]{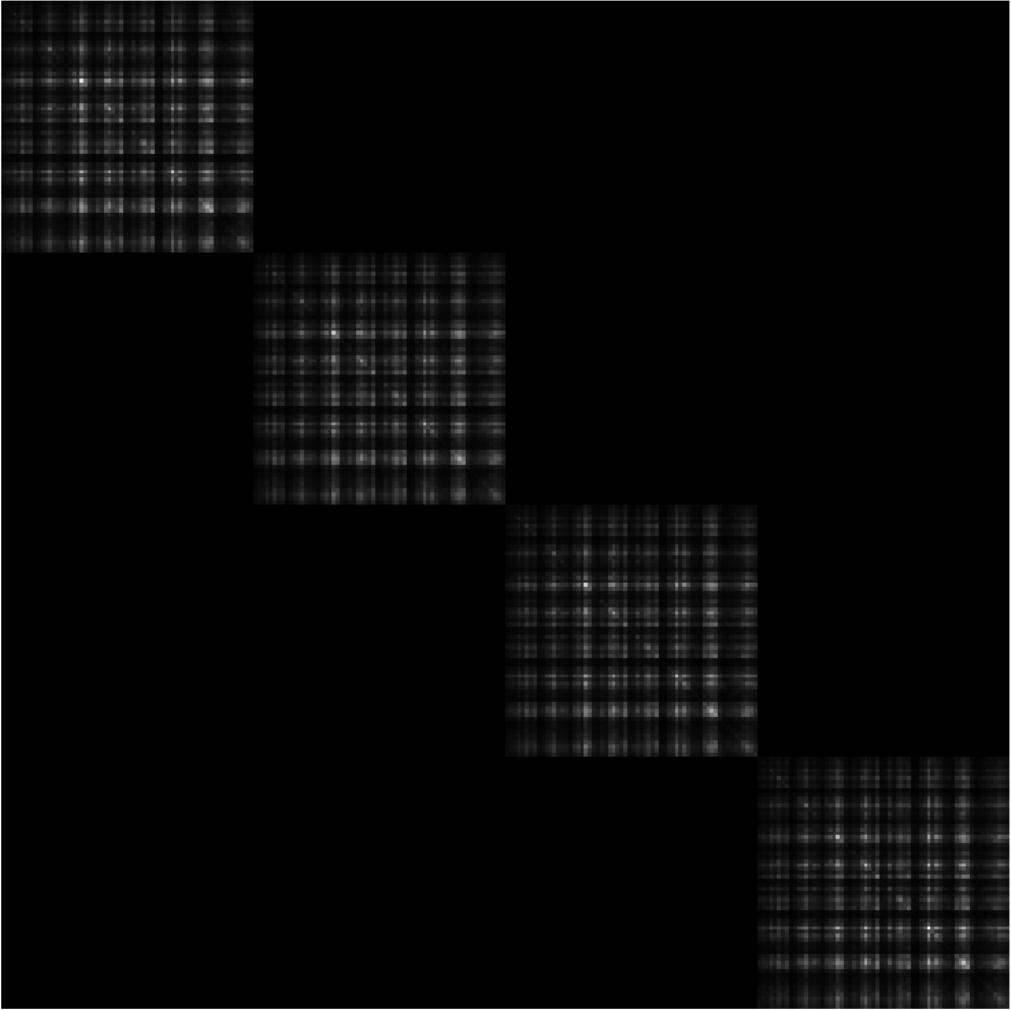}
  \caption{
  Example of $V^{t}_{\vx^\prime\vx^\prime}$ for batch size $B=4$ in DIGITS dataset.
  Higher (whiter) values concentrate along the diagonal blocks $V^{t}_{\vx^{(i)}\vx^{(i)}}$
  }
  \label{fig:Vxx-avg}
  \vskip -0.2in
\end{wrapfigure}
\textbf{Block-diagonal Value Hessian:}
Extending the Bellman optimization framework to accept mini-batch samples $\{\vx_0^{(i)}\}_{i=0}^{B}$
has been made in previous work \cite{liu2020differential} by
augmenting the state space to $\vx_t^\prime = [\cdots,\vx_t^{(i)},\cdots]^\transpose$.
However, such a formulation can cause memory explosion
when $\vx_t$ is lifted to 3D feature map in convolutional layers,
let alone the augmented value function considered in GT-DDP (cf \eq{\ref{eq:values}}).
In this work, we propose to approximate the batch-augmented value Hessian $V^t_{\vx^\prime\vx^\prime}$ as block-diagonal.
The approximation is made from an empirical observation (see Fig.~\ref{fig:Vxx-avg}) that
$V^{t}_{\vx^\prime\vx^\prime}$ contains only nontrivial values along the diagonal blocks,
 even when networks contain Batch Normalization (BN) layers.
This suggests one can reduce the memory consumption by approximating the batch-augmented value Hessian as block-diagonal and only
carry batch matrices, $\{V^{t}_{\vx^{(i)}\vx^{(i)}}\}_{i=0}^{B}$, along the backward computation.

\textbf{Gauss-Newton (GN) Approximation at the Terminal Hessian:}
Next, \\
we impose Gauss-Newton approximation to the Hessian at the prediction \\
layer.
Surprisingly, this will lead to a nontrivial factorization in the Bellman optimization framework.
For dynamics represented by feedforward networks, we have the following proposition.
\begin{proposition}[Outer-product factorization in DDP] \label{prop:gn-ddp}
Consider the following form of OCP:
\begin{equation}
\min_{\vuTraj}
\left[
    \phi(\vx_{T}) + \sum_{t=0}^{T-1}\ell_t(\vu_{t})
\right]
\quad \text{s.t. } \vx_{t+1} = f_t(\vx_{t}, \vu_{t}) \period
\label{eq:ocp2}
\end{equation}
If the Hessian of the terminal loss can be expressed by an outer product of vectors,
i.e. $\nabla^2\phi(\vx_{T}) \approx \vz_\vx^T \otimes \vz_\vx^T$ for some vector $\vz_\vx^T$ (e.g. $\vz_\vx^T=\nabla \phi$ for GN approximation), then we have the factorization:
\begin{equation}
\begin{split}
\forall t \comma \quad
\Quxt = \vq_\vu^t \otimes \vq_\vx^t \comma \quad
\Qxxt = \vq_\vx^t \otimes \vq_\vx^t \comma \quad
V_{\vx\vx}^t = \vz_\vx^t \otimes \vz_\vx^t \comma \quad
\end{split} \label{eq:qxqu}
\end{equation}
where $\vq_\vu^t$, $\vq_\vx^t$, and $\vz_\vx^t$
are outer-product vectors
which can be computed backward:
\begin{align}
\vq_\vu^t = \futT \vz_\vx^{t+1} \comma \quad
\vq_\vx^t = \fxtT \vz_\vx^{t+1} \comma \quad
\vz_\vx^t = \sqrt{1+\vq_\vu^\ttranspose \QuuInvt \vq_\vu^t } \text{ }\vq_\vx^t
\period
\label{eq:vxvx}
\end{align}
\end{proposition}
In other words,
the outer-product factorization at the final stage can be backward propagated to all proceeding layers.
Thus, \textit{state-dependent} second-order matrices can be represented as outer products of vectors.
We note that the low-rank structure at the prediction layer has been observed
when classification loss (e.g. cross-entropy) is used \cite{nar2019cross,lezama2018ole}.
Prop. \ref{prop:gn-ddp} can be extended to residual networks:
\begin{proposition}[Outer-product factorization in GT-DDP]
The residual value Hessians considered in \eq{\ref{eq:vxx_res}},
when the same outer-product factorization is imposed at the terminal stage,
take the form
\begin{equation}
\begin{split}
V^{t}_{\vx\vxr} = \vz_\vx^t \otimes \vz_\vxr^t
\text{ } \text{ and } \text{ }
V^{t}_{\vxr\vxr} = \vz_\vxr^t \otimes \vz_\vxr^t \comma
\text{ } \text{ where }
\vz_\vxr^t = \sqrt{1+\vq_\vu^\ttranspose \QuuInvt \vq_\vu^t } \text{ }\vz_\vxr^{t+1}
\end{split} %
\end{equation}
and $(\vq_\vu^t,\vq_\vx^t,\vz_\vx^t)$ are given by Eq. (\ref{eq:vxvx}).
When the non-identity shortcut mapping, i.e. $h_t(\vx_r,\vv_t)$ in \eq{\ref{eq:coop-traj}}, is presented,
the cooperative forms of $\vz_\vx^t$ and $\vz_\vxr^t$, denoted $\tilde{\vz}_\vx^t$ and $\tilde{\vz}_\vxr^t$,
are given by
\begin{equation}
\begin{split}
\tilde{\vz}_\vx^t = \sqrt{1+\vq_\vu^\ttranspose \QuuInv \vq_\vu^t + \vq_\vv^\ttranspose \QvvInv \vq_\vv^t} \text{ }\vq_\vx^t
\comma \quad
\tilde{\vz}_\vxr^t = \sqrt{1+\vq_\vu^\ttranspose \QuuInv \vq_\vu^t + \vq_\vv^\ttranspose \QvvInv \vq_\vv^t} \text{ }\vq_\vxr^t
,
\end{split} %
\end{equation}
where
$\vq_\vxr^t=h_{\vxr}^\ttranspose \vz_\vxr^{t+1}$,
and $\vq_\vv^t=h_{\vv}^\ttranspose   \vz_\vxr^{t+1}$.
\end{proposition}

The outer-product factorization, together with the block-diagonal approximation, reduces the computational dependency
by dropping the memory by $2/3$ and the runtime by $1/5$ compared with previous work \cite{liu2020differential},
as shown in Fig.~\ref{fig:ddp-compare}.
As such, we adopt both approximation in all experiments.

\textbf{Jacobian of Layers Dynamics:}
Finally, computing the derivatives of the Bellman objective involve evaluating
the Jacobian associated with each layer,
e.g.
$Q^t_{\vx}={f^t_{\vx}}^\transpose V^{t+1}_{\vx}$ and
$Q^t_{\vu}={f^t_{\vu}}^\transpose V^{t+1}_{\vx}$.
These computations can be done efficiently for
both fully-connected (FC) and
convolution (Conv) layers:
\begin{align*}
{f^t_{\vx}}^\transpose V^{t+1}_{\vx}=\Big\{
\begin{array}{l}
  {\mW_t^\transpose V^{t}_{\vh}} \\
  {\omega_t^\transpose \ConvT V^{t}_{\vh}}
\end{array}
\comma \quad
{f^t_{\vx}}^\transpose V^{t+1}_{\vx}=\Big\{
\begin{array}{l}
  {\vx_t \otimes V^{t}_{\vh}} \\
  {\vx_t \ConvT V^{t}_{\vh}}
\end{array}
\comma \quad \text{where }
f^t_{\vx}=\sigma_t(\vh_t)
\comma \quad
\vh_t \triangleq \Big\{
\begin{array}{l}
  {\mW_t\vx_t+\vb_t} \\
  {\omega_t * \vx_t}
\end{array}
\end{align*}
respectively denote the pre-activation of FC and Conv layers.
$*$ and $\ConvT$ denote the convolution and deconvolution (transposed convolution) operator
\cite{dumoulin2016guide,zeiler2010deconvolutional}.

\section{Evaluation on Classification Data Set} \label{sec:experiment}

\newcommand{\specialcell}[2][c]{%
  \begin{tabular}[#1]{@{}c@{}}#2\end{tabular}}

\newcommand{\bestTest}[1]{{ (\textbf{$\textbf{#1}$}) }}

\def\Plus{\texttt{+}}
\def\Minus{\texttt{-}}
\def\win{{\markgreen{\text{ (\Plus)}}}}
\def\lose{{\markredd{\text{ (\Minus)}}}}

\bgroup
\setlength\tabcolsep{0.04in}
\begin{table*}[h]
\vskip -0.2in
\captionsetup{justification=centering}
\caption{Performance comparison on train loss and validation accuracy (over $6$ random seeds). \\ \win $\text{ }$ and \lose $\text{ }$ respectively denote improvement and degradation over non-GT-DDP baselines.
}
\label{table:training}
\vskip -1.5in
\begin{center}
\fontsize{8.5}{11}\selectfont{
\begin{tabular}{c|r?cccc|cccc}
\toprule
\multicolumn{1}{c}{} & Data Set
& SGD & RMSProp & Adam & EKFAC & \specialcell{GT-DDP\\-SGD} & \specialcell{GT-DDP\\-RMSProp} & \specialcell{GT-DDP\\-Adam} & \specialcell{GT-DDP\\-EKFAC} \\
\midrule
\multirow{5}{*}{\rotatebox[origin=c]{90}{Training}}&
DIGITS&
$0.0053$&$0.0247$&$0.0182$&$0.0514$&$\textbf{0.0050}$\win &$0.0124$\win &$0.0081$\win &$0.0514$\win \\
&MNIST&
$0.0250$&$0.0284$&$0.0330$&$0.0290$&$\textbf{0.0240}$\win&$0.0282$\win&$0.0312$\win&$0.0291$\lose\\[1pt]
&SVHN&
$0.2755$&$0.2670$&${0.2544}$&${0.2049}$&${0.2692}$\win&$0.2637$\win&${0.2517}$\win&$\textbf{0.2047}$\win\\[1pt]
&CIFAR-10&
$0.0296$&$0.0107$&$0.0127$&$0.0922$&${0.0284}$\win&$\textbf{0.0069}$\win&$0.0096$\win&$0.0907$\win\\[1pt]
&CIFAR-100&
$0.0075$&$0.0058$&$0.0055$&$0.0120$&${0.0075}$\lose&${0.0058}$\win&$\textbf{0.0054}$\win&$0.0125$\lose\\[1pt]
\midrule
\multirow{5}{*}{\rotatebox[origin=c]{90}{Validation (\%)}}&
DIGITS&
$96.09$&$95.61$&$95.81$&$95.31$&$\textbf{96.10}$\win&$95.92$\win&$95.84$\win&$95.55$\win\\[1pt]
&MNIST&
$98.59$&$98.52$&$98.51$&$98.56$&$\textbf{98.62}$\win&$98.53$\win&$98.51$\win&$98.56$\lose\\[1pt]
&SVHN&
$88.58$&$88.96$&${89.20}$&$88.75$&${89.90}$\win&$89.02$\win&$\textbf{89.22}$\win&$89.91$\win\\[1pt]
&CIFAR-10&
${74.69}$&$70.88$&$72.51$&$74.33$&$\textbf{74.69}$\win&$70.97$\win&$72.68$\win&$74.18$\lose\\[1pt]
&CIFAR-100&
$71.78$&$71.65$&$71.96$&$71.95$&${72.06}$\win&$71.91$\win&$72.19$\win&$\textbf{72.24}$\win\\[1pt]
\bottomrule
\end{tabular}
}
\end{center}
\vskip -0.05in
\end{table*}
\egroup

\begin{wrapfigure}{r}{0.2\textwidth}
  \vskip -0.21in
    \includegraphics[width=0.2\textwidth]{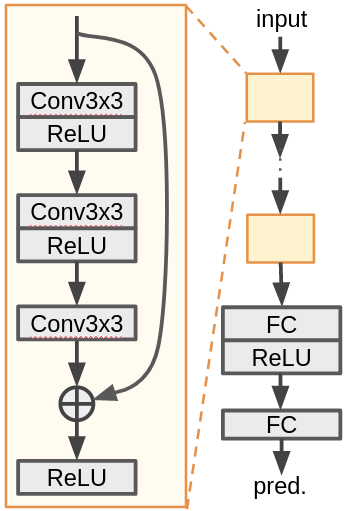}
  \caption{
  Architecture and residual block in Sec. \ref{sec:experiment}.
  }
  \label{fig:architecture}
  \vskip -0.2in
\end{wrapfigure}
In this section we verify the performance of our GT-DDP optimizer and discuss the benefit of having layer-wise feedback policies during weight update.
Detail experiment setup and additional results are provided in the Appendix \ref{app:exp}.

We validate the performance of GT-DDP on
digits recognition and image classification data set.
The networks consist of $1$-$4$ residual blocks followed by fully-connected (FC) layers (see Fig.~\ref{fig:architecture}),
except that we use ResNet18 \cite{he2016deep} for the CIFAR-100 dataset.
Each block contains a skip connection between $3$ convolution modules, possibly with a non-identity shortcut mapping if needed.
Following the discussion in the previous section,
we select our baselines as SGD, RMSprop \cite{hinton2012neural}, Adam \cite{kingma2014adam}, and EKFAC \cite{george2018fast},
as they cover most widely-used curvature approximation in training deep nets,
including (adaptive) diagonal matrices and second-order Kronecker factorization.

Table~\ref{table:training} summarizes our main results.
In each experiment we keep the shared hyper-parameters (e.g.
learning rate and weight decay) between baselines and their GT-DDP variants the same,
so that the performance difference only comes from GT-DDP framework.
On all data set, GT-DDP achieves better or comparable results on both training and accuracy.
Notably, when comparing original methods with their GT-DDP integrated variants,
the latter improve training convergence on {almost all} dataset.
Empirically, it also leads to better generalization.

\begin{figure}[t]%
  \centering
  \vskip -0.25in
  \subfloat{
  	\includegraphics[width=0.50\linewidth]{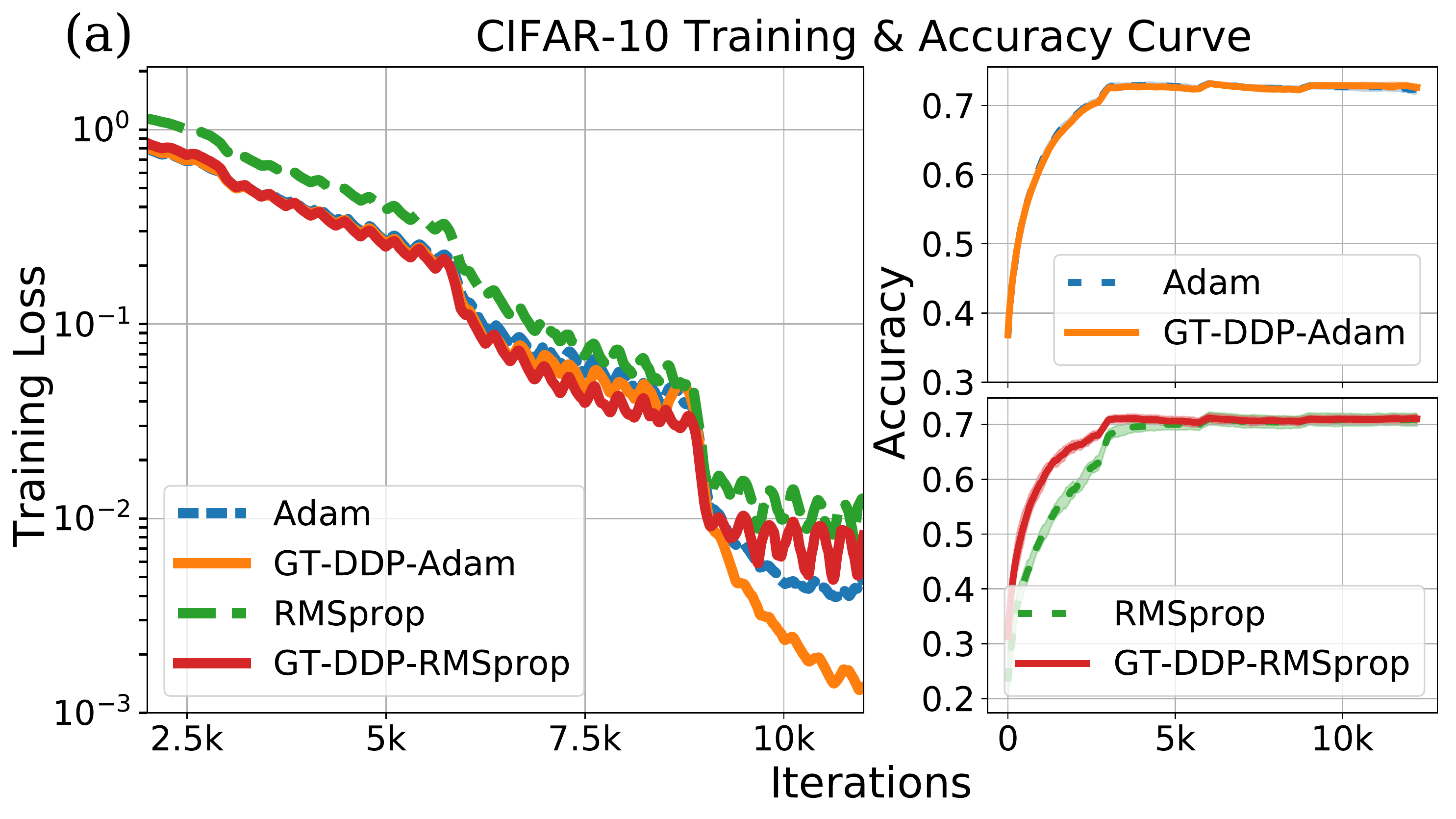}
  	\label{fig:cifar-100-train}
  	}
  \subfloat{
  	\includegraphics[width=0.28\linewidth]{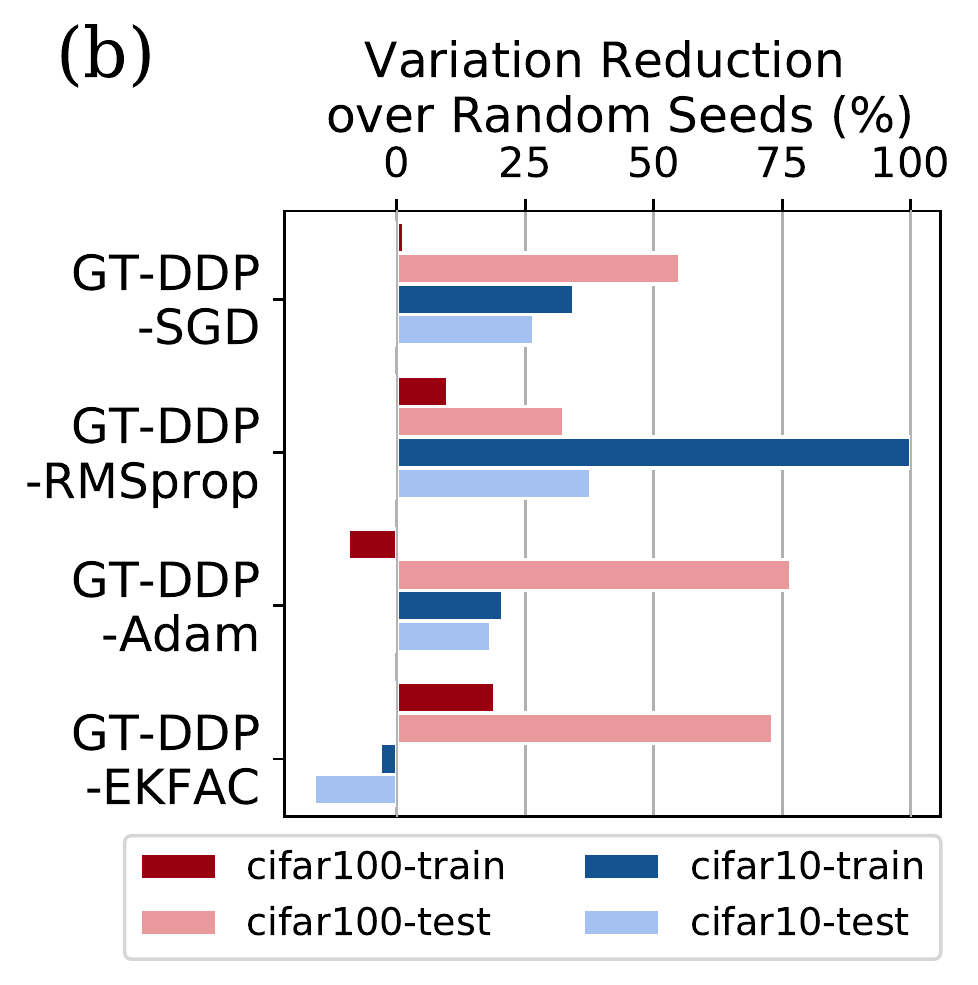}
  	\label{fig:cifar-100-var}
  	}
  \subfloat{
  	\includegraphics[width=0.2\linewidth]{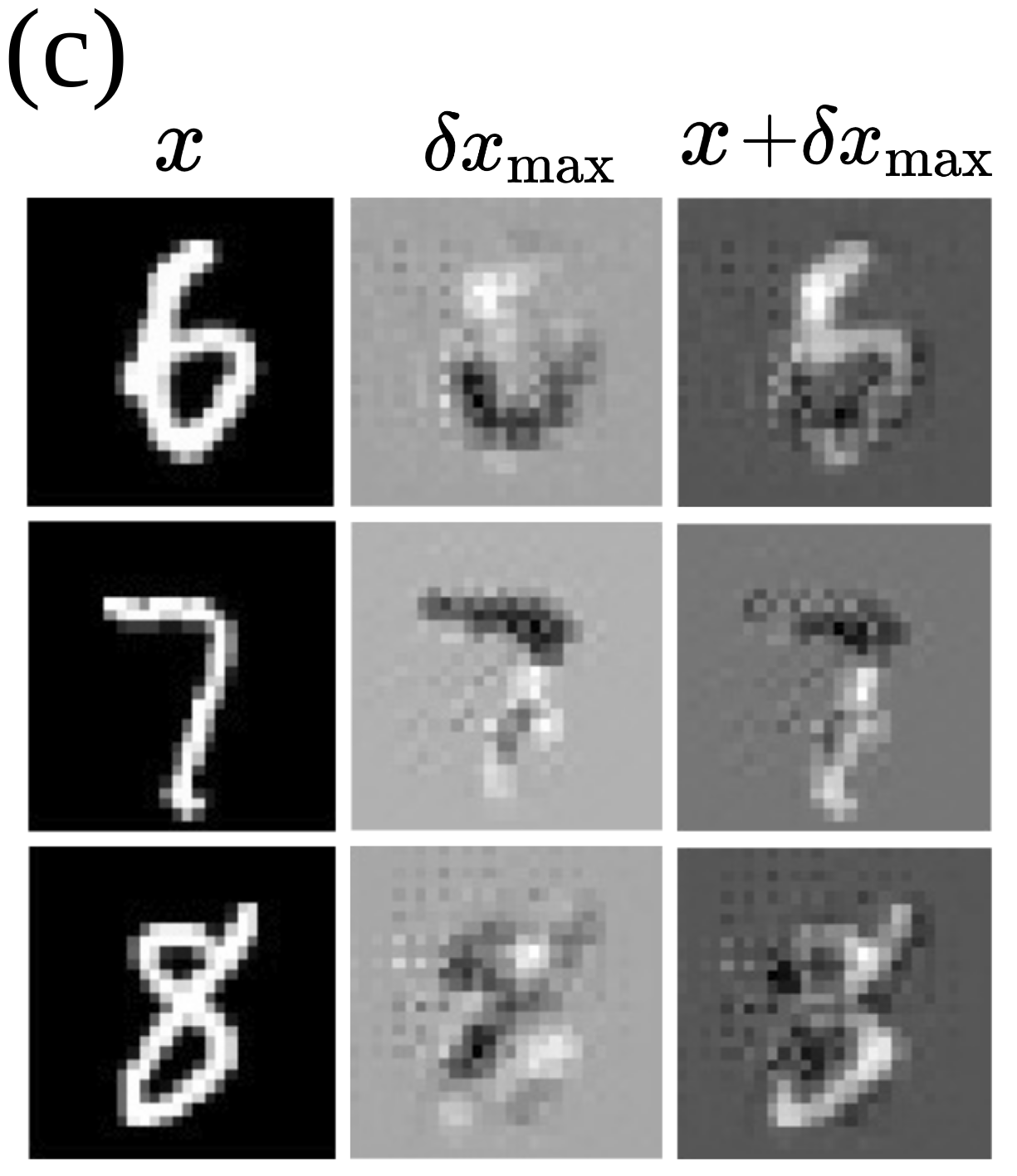}
  	\label{fig:K-vis}
  	}
  \vskip -0.05in
  \caption{
    (a) Training performance on CIFAR-10 for Adam, RMSprop and their GT-DDP variants.
    (b) Variation reduction over $3$-$6$ random seeds
     on CIFAR-10 and CIFAR-100.
      We report the value $(\Var_{\text{GT-DDP-Baseline}}-\Var_{\text{Baseline}})/\Var_{\text{Baseline}}$. %
    (c) Visualization of the feedback policies on MNIST.
  }%
  \label{table:fig:2}%
  \vskip -0.15in
\end{figure}

Since the feedback updates
are typically order of magnitude smaller than the open gain due to the sparse Hessian of standard classification loss (i.e. cross-entropy),
GT-DDP follows similar training trend with the baseline it used to approximate the parameter curvature (see Fig.~\ref{fig:cifar-100-train}).
Nevertheless,
these additional updates have a non-trivial effect on not only improving the convergence
but robustifying the training.
As shown in Fig.~\ref{fig:cifar-100-var},
GT-DDP reduces the variation of the performance difference over random seeds subjected to same hyper-parameters\footnote{
Additional experiments across different hyper-parameters are provided in the Appendix \ref{app:exp}.}.
In fact,
the Bellman framework has been shown
numerically stable than direct optimization such as Newton method \cite{liao1992advantages},
since it takes into account the temporal, i.e. layer-wise, structure inherit in \eq{\ref{eq:ocp}}.
As the concern for reproducibility arises \cite{henderson2018deep},
GT-DDP provides a principled way to improve the robustness and consistency during training.
We highlight this perspective as the benefit gained from \textit{architecture-aware} optimization.

To understand the effect of feedback policies more perceptually,
we conduct eigen-decomposition on the feedback matrices of convolution layers and project the leading eigenvectors back to image space, using techniques proposed in \cite{zeiler2014visualizing}.
These feature maps, denoted $\delta x_{\max}$ in Fig.~\ref{fig:K-vis}, correspond to the dominating differential image that GT-DDP policies shall respond with during weight update.
Fig.~\ref{fig:K-vis} demonstrates that the feedback policies indeed capture
non-trivial visual feature related to the pixel-wise difference between spatially similar classes, e.g. $(8,3)$ or $(7,1)$.
We note that these differential maps differ from adversarial perturbation \cite{goodfellow2014explaining}
as the former directly link the parameter update to the change in activation;
thus being more interpretable.

\section{Discussion on Game-Theoretic Second-order Optimizer} \label{sec:discussion}
\begin{figure}[h]%
  \vskip -0.33in
  \subfloat{\includegraphics[width=0.3\linewidth]{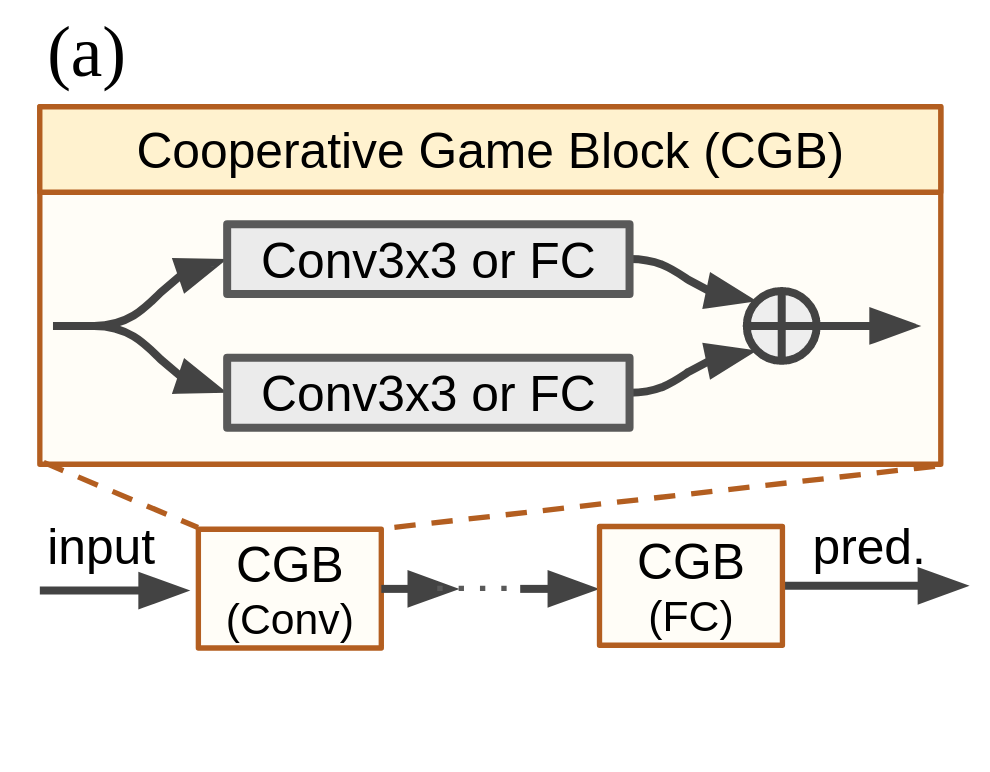}
  	\label{fig:cg-net}
  }
  \subfloat{\includegraphics[width=0.7\linewidth]{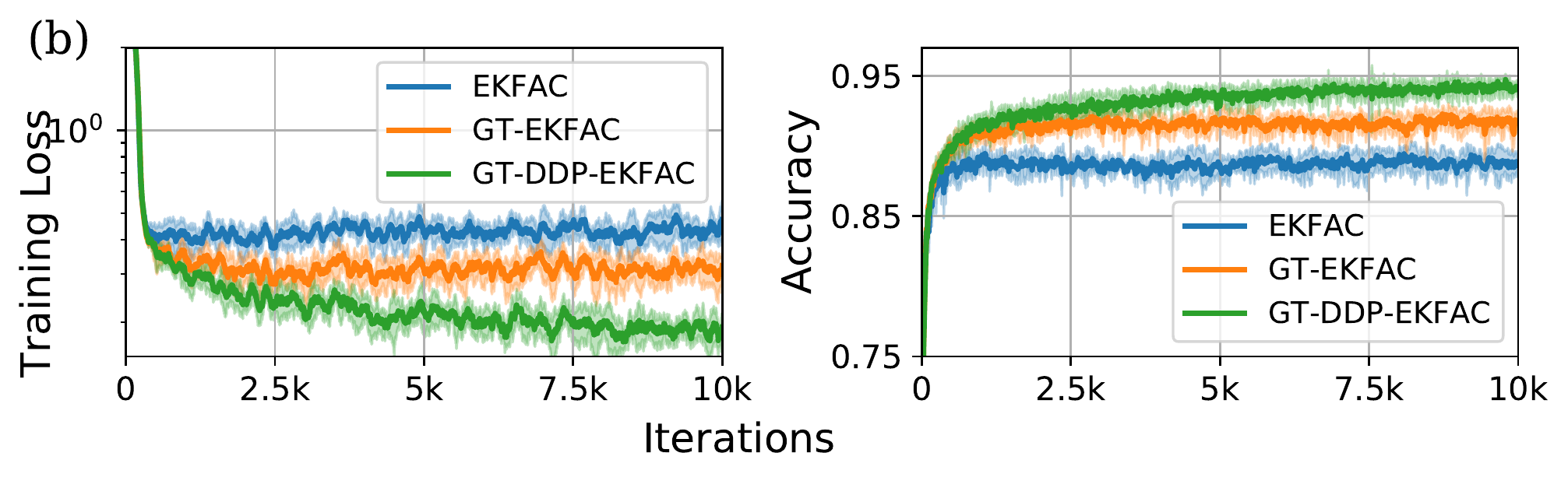}
  	\label{fig:gt-ekfac}
  }
  \vskip -0.17in
  \caption{
    (a) Illustration of the cooperative-game module.
    (b) Training and testing performance on MNIST using the architecture in \ref{fig:cg-net}.
    GT-EKFAC denotes integration of EKFAC with Corollary \ref{coro:5}.\\
  }%
  \label{table:fig:2}%
  \vskip -0.15in
\end{figure}

Theorem \ref{thm:coop} may be of independent interest for developing game-theoretic second-order optimizer,
as Eq.(\ref{eq:thm1},\ref{eq:thm2}) provide efficient second-order approximation to the cooperative Hessian
regardless of the presence of Bellman framework.
To better show its effectiveness, let us consider the modules in Fig.~\ref{fig:cg-net}
that resemble the cooperative game,
i.e. two (p)layers take the same input and affect each payoff through output addition.
Such an architecture has also appeared in recent work of progressive training \cite{wu2019splitting}.
Interestingly, for this particular structure,
we have the following corollary to Thm. \ref{thm:coop}:
\begin{corollary} \label{coro:5}
Let
$Q_{\vu\vu} \approx \Auu \otimes \Buu = \mU\Sigma_{\vu\vu}\mU^\transpose$
be the eigen-decomposition of the Kronecker factorization,
where
$\Sigma_{\vu\vu}=\diag(\lambda_{\vu\vu})+\gamma\mI$ and
$\gamma>0$ is the Tikhonov damping.
Consider the architecture in Fig.~\ref{fig:cg-net},
its cooperative matrix corresponds to rescaling in the eigenspace of $\Quu$, i.e.
\begin{align}
\QuuC=\mU \tilde{\Sigma}_{\vu\vu} \mU^\transpose \comma \quad
\tilde{\Sigma}_{\vu\vu} = \diag(\tilde{\lambda}_{\vu\vu})+\gamma\mI \comma \quad \text{and } \quad
	\tilde{\lambda}_{\vu\vu}^i = \frac{\gamma}{\gamma+{\lambda}_{\vu\vu}^i} {\lambda}_{\vu\vu}^i
\period
\end{align}
\end{corollary}
Notice that $\frac{\gamma}{\gamma+{\lambda}_{\vu\vu}^i}\le1$ for positive eigenvalues;
thus
the inverse Hessian $\QuuCInv$ shall take a larger step in eigenspace compared with $\QuuInv$.
As shown in Fig~\ref{fig:gt-ekfac}, integrating this game theoretic perspective with existing second-order methods,
denoted GT-EKFAC, leads to better convergence.
Having additional layer-wise policies from the GT-DDP framework further improves the performance.

\section{Conclusion}
In this work, we present the Game-Theoretic Differential Dynamic Programming (GT-DDP) optimizer
as a new class of second-order algorithm.
Theoretically,
we strengthen the optimal control connection proposed in previous work by showing
training residual networks can be linked to trajectory optimization in a cooperative game.
Algorithmically,
we propose several effective approximation which scales GT-DDP to training modern architectures
and suggest how existing methods can be extended to accept such a game-theoretic perspective.
We validate GT-DDP on several image classification dataset,
showing improvement on both convergence and robustness.

\newpage

\newpage

\section*{Acknowledgments}
The authors would like to thank Chen-Hsuan Lin, Yunpeng Pan, and Yen-Cheng Liu for many helpful discussions on the paper.
The work is supported under Amazon AWS Machine Learning Research Award (MLRA).

\bibliographystyle{unsrtnat}
\bibliography{reference.bib}

\newpage
\appendix
\begin{center}
\textbf{{\huge Supplementary Material}}
\end{center}

\section{Derivation of Optimal Control Representation for Residual networks} \label{app:sec3-dev}

\subsection{Derivation in Section \ref{sec:31}} \label{app:sec31-dev}

First, recall the Bellman objective in feedforward networks, 
$Q_t(\vx_t,\vu_t) \triangleq \ell_t(\vu_t) + V_{t+1}(f_t(\vx_t,\vu_t))$.
Following standard chain rule, the second-order expansion of $Q_t$ in \eq{\ref{eq:du-star-ddp}} takes the form
\arraycolsep=1.4pt\def\arraystretch{1.5}
\begin{align} \label{eq:q-matrix}
\delta Q_t =
\frac{1}{2}
\left[\begin{array}{c}
	\mathbf{1} \\ {\delta \vx_t} \\ {\delta \vu_t}
\end{array}\right]^{\transpose}
\left[\begin{array}{ccc}
	\mathbf{0} & {\Qxt^{\transpose}} & {\Qut^{\transpose}} \\ 
	{\Qxt} & {\Qxxt} & {\Qxut} \\ 
	{\Qut} & {\Quxt} & {\Quut}
\end{array}\right]
\left[\begin{array}{c}
	\mathbf{1} \\ {\delta \vx_t} \\ {\delta \vu_t}
\end{array}\right]
\text{, }
\begin{array}{r@{=}}
{\Qxt \text{ }\text{ }} \\
{\Qut \text{ }\text{ }} \\
{\Quut } \\
{\Quxt } \\
{\Qxxt } 
\end{array}
\begin{array}{l@{}}
{\fxtT \Vxt} \\
{\futT \Vxt + \lut} \\
{\futT \Vxxt \fut+\Vxt \cdot \fuut + \luut} \\
{\futT \Vxxt \fxt+\Vxt \cdot \fuxt} \\
{\fxtT \Vxxt \fxt+\Vxt \cdot \fxxt } 
\end{array}
\comma
\end{align}
\arraycolsep=1.4pt\def\arraystretch{1.0}
where the dot notation represents the product of a vector with a 3D tensor.
Note that in practice, 
the dynamics is often expanded up to the first order, i.e. by omitting $\fxxt$,$\fuut$,$\fuxt$ above,
while keeping the full second-order expansion of the value function $\Vxxt$.
This can be seen as Gauss-Newton (GN) approximation, and the stability obtained by keeping only the linearized dynamics is discussed thoroughly in trajectory optimization \cite{tassa2012synthesis,todorov2005generalized}.
As such, both DDP \cite{liu2020differential} and our GT-DDP optimizer adopt the same setup.

\def\vX{{\hat{\vx}}}
\def\dvX{{\delta \hat{\vx}}}

\def\VXt{{\hat{V}^{t+1}_{\vX}}}
\def\VXXt{{\hat{V}^{t+1}_{\vX\vX}}}

\def\hfxt{{\hat{f}^t_{\vX}}}
\def\hfut{{\hat{f}^t_{\vu}}}

\def\hfu{{\hat{f}_{\vu}}}
\def\hfuT{{\hat{f}_{\vu}^\transpose}}
\def\hfxT{{\hat{f}_{\vX}^\transpose}}
\def\hfutT{{\hat{f}_{\vu}^{t \text{ }\transpose}}}
\def\hfxtT{{\hat{f}_{\vX}^{t \text{ }\transpose}}}
\def\hfxx{{\hat{f}_{\vX \vX}}}
\def\hfuu{{\hat{f}_{\vu \vu}}}
\def\hfux{{\hat{f}_{\vu \vX}}}
\def\hfxu{{\hat{f}_{\vX \vu}}}
\def\zero{{\mathbf{0}}}

\def\hQxt{{\hat{Q}^t_{\vX}}}
\def\hQut{{\hat{Q}^t_{\vu}}}
\def\hQxxt{{\hat{Q}^t_{\vX \vX}}}
\def\hQuxt{{\hat{Q}^t_{\vu \vX}}}
\def\hQxut{{\hat{Q}^t_{\vX \vu}}}
\def\hQuut{{\hat{Q}^t_{\vu \vu}}}

\def\hQxts{{\hat{Q}^{t_s}_{\vX}}}
\def\hQuts{{\hat{Q}^{t_s}_{\vu}}}
\def\hQuuts{{\hat{Q}^{t_s}_{\vu \vu}}}
\def\hQuxts{{\hat{Q}^{t_s}_{\vu \vX}}}
\def\hQxuts{{\hat{Q}^{t_s}_{\vX \vu}}}
\def\hQxxts{{\hat{Q}^{t_s}_{\vX \vX}}}
\def\kts{{\mathbf{k}_{t_s}}}
\def\Kts{{\mathbf{K}_{t_s}}}
\def\Gts{{\mathbf{G}_{t_s}}}
\def\KtsT{{\mathbf{K}_{t_s}^\transpose}}
\def\GtsT{{\mathbf{G}_{t_s}^\transpose}}

Now, let us consider the value minimization described in \eq{\ref{eq:values}} for residual networks.
We shall interpret the propagation rules as $\hat{f}_t(\vX_t,\vu_t)$,
where $\vX_t$ is the \textit{residual-augmented} state $\vX_t \triangleq [\vx_t,\vx_r]^\transpose$.
The Jacobian of this state-augmented dynamics and its relation to the ones in the absence of residual paths, i.e. $\fut$ $\fxt$, 
can be summarized below:
\begin{subequations}
\begin{alignat}{3}
\text{At $t = t_f$, \eq{\ref{eq:value1}} gives }
\underbrace{
	\vx_{t+1} 
		= \vx_r + f_{t}(\vx_t,\vu_t)
	}_{\vx_{t+1} = \hat{f}_t(\vX_t,\vu_t)}
\Rightarrow
	\hfxt &= \left[\fxt \quad \mI\right] &&\comma \text{ } 
	\hfut = \fut \comma \label{eq:dynamics1}
\\
\text{At $t \in (t_s,t_f)$, \eq{\ref{eq:value2}} gives }
\underbrace{
	\left[\begin{array}{c} {\vx_{t+1}} \\ {\vx_r} \end{array}\right]
		= \left[\begin{array}{c} {f_{t}(\vx_t,\vu_t)} \\ {\vx_r} \end{array}\right]
	}_{\vX_{t+1} = \hat{f}_t(\vX_t,\vu_t)}
\Rightarrow
	\hfxt &= 
		\left[\begin{array}{cc} {\fxt} & {\zero} \\ {\zero} & {\mI} \end{array}\right] &&\comma \text{ }
	\hfut = \left[\begin{array}{c} {\fut} \\ {\zero} \end{array}\right] \comma \label{eq:dynamics2}
\\
\text{At $t = t_s$, \eq{\ref{eq:value3}} gives}
\underbrace{
	\left[\begin{array}{c} {\vx_{t+1}} \\ {\vx_r} \end{array}\right]
		= \left[\begin{array}{c} {f_{t}(\vx_t,\vu_t)} \\ {\vx_t} \end{array}\right]
	}_{\vX_{t+1} = \hat{f}_t(\vx_t,\vu_t)}
\Rightarrow
	\hfxt &= \left[\begin{array}{c} {\fxt} \\ {\mI} \end{array}\right]  &&\comma \text{ } 
	\hfut  = \left[\begin{array}{c} {\fut} \\ {\zero} \end{array}\right] \comma \label{eq:dynamics3}
\end{alignat} \label{eq:dynamics-res}
\end{subequations}
where $\mI$ is the identity matrix.

Once we have the explicit form of dynamics written, 
the optimal control representation can be derived by substituting \eq{\ref{eq:dynamics-res}} into \eq{\ref{eq:q-matrix}}.
After some algebra, one can verify that for $t\in(t_s,t_f]$ we will have
\begin{subequations}
\begin{align}
\hQxt &= \hfxtT \VXt =
\left[ {\Qxt} \quad {V^{t+1}_{\vx_r}} \right]^\transpose \\
\hQut &= \hfutT \VXt + \lut = \Qut\\
\hQuut &=\hfutT \VXXt \hfut + \luut = \Quut \\
\hQuxt &=\hfutT \VXXt \hfxt = \left[ \Quxt \quad \futT V^{t+1}_{\vx\vx_r} \right] \\
\hQxxt &=\hfxtT \VXXt \hfxt
 = \left[\begin{array}{cc} {\Qxxt} & {\fxtT V^{t+1}_{\vx \vx_r}} \\ {V^{t+1}_{\vx_r \vx}\fxt} & {V^{t+1}_{\vx_r\vx_r}} \end{array}\right] \period
\end{align}
\end{subequations}
The optimal feedback policy is given by
\begin{align}
    \dvu_t^* (\dvX_t) = - (\hQuut)^{\Inv}(\hQut+\hQuxt\dvX_t) =
    \kt + \Kt \dvx_t \underbrace{-(\Quut)^\Inv {f^t_{\vu}}^\transpose V^{t+1}_{\vx\vxr}}_{\triangleq \Gt} \dvx_r
     \comma
    \label{eq:du-res-app}
\end{align}
Note that 
$\VXt$ and $\VXXt$ are the derivatives of the value function $\hat{V}_{t+1}(\vX_{t+1})$ induced by the state-augmented dynamics.
We can compute these matrices backward from $t_f$ similar to \eq{\ref{eq:value-recursive}}:
\begin{alignat*}{3}
\hat{V}^{t}_{\vX}
     &= \hQxt - \hQxut (\hQuut)^{\Inv} \hQut
     &&= \left[\begin{array}{c} {\Qxt + \Qxut \kt} \\ {V^{t+1}_{\vx_r} - \GtT\Quut\kt} \end{array}\right]
     \triangleq \left[\begin{array}{c} {V^t_{\vx}} \\ {V^t_{\vx_r}} \end{array}\right] \numberthis \comma \label{eq:vx_res_app}
     \\
\hat{V}^{t}_{\vX\vX}
     &=  \hQxxt - \hQxut (\hQuut)^{\Inv} \hQuxt
     &&= \left[\begin{array}{cc} {\Qxxt + \Qxut \Kt} & { {f^t_{\vx}}^\transpose V^{t+1}_{\vx\vxr} - \KtT \Quut \Gt } \\
                               { V^{t+1}_{\vxr\vx}f^t_{\vx} - \GtT \Quut \Kt } & { V^{t+1}_{\vxr\vxr} - \GtT \Quut \Gt } \end{array}\right] \\
&\quad
     &&\triangleq \left[\begin{array}{cc} {V^t_{\vx\vx}} & {V^t_{\vx\vx_r}} \\
                                          {V^t_{\vx_r\vx}} & {V^t_{\vx_r\vx_r}} \end{array}\right] \numberthis \comma \label{eq:vxx_res_app}
\end{alignat*}
with the terminal conditions given by
$V^{t_f+1}_{\vxr}=V^{t_f+1}_{\vx}$ and
$V^{t_f+1}_{\vxr\vxr}=V^{t_f+1}_{\vx\vxr}=V^{t_f+1}_{\vxr\vx}=V^{t_f+1}_{\vx\vx}$.

As for the stage $t=t_s$ where the residual state is split out,
the derivatives of $Q_t$ follow (again one can readily verify by substituting \eq{\ref{eq:dynamics3}} into \eq{\ref{eq:q-matrix}}) by
\begin{subequations}
\begin{align}
\hQxts &= \Qxts +V^{t_s+1}_{\vx_r} \comma \\
\hQuxts &=\Quxts + \futsT V^{t_s+1}_{\vx\vx_r} \comma \\
\hQxxts &= \Qxxts + \fxtsT V^{t_s+1}_{\vx \vx_r} + V^{t_s+1}_{\vx_r \vx}\fxt + V^{t_s+1}_{\vx_r\vx_r} \comma
\end{align}
\end{subequations}
and $\hQut$ and $\hQuut$ remain the same with $\Qut$ and $\Quut$.
The resulting optimal policy admits the same form as in \eq{\ref{eq:du-res-app}}.

\textbf{Proof of Proposition \ref{prop:res}:}
Finally, one can verify
Eq.~(\ref{eq:vx-res}, \ref{eq:vxx-res2}) by noticing that the derivatives of the value function at $t=t_s$ follow
\begin{align*} 
\tilde{V}^{t_s}_{\vx} 
     &= \hQxts - \hQxuts (\hQuuts)^{\Inv} \hQuts \\
     &= (\Qxts + \Qxuts \kts) + (V^{t_s+1}_{\vx_r} - \futsT V^{t_s+1}_{\vx\vx_r} \kts) \\
     &= V^{t_s}_{\vx}+ V^{t_s+1}_{\vx_r} - \GtsT\Quuts\kts \\
     &= V^{t_s}_{\vx}+ V^{t_f+1}_{\vx} - \textstyle \sum_{t \in [t_s, t_f]} \GtT\Quut\kt \comma \numberthis \label{eq:vx-res-app} \\
\tilde{V}^{t_s}_{\vx\vx} 
     &= \hQxxts - \hQxuts (\hQuuts)^{\Inv} \hQuxts \\
     &= (\Qxxts + \Qxuts \Kts) + (\fxtsT V^{t_s+1}_{\vx \vx_r} - \KtsT \Quuts \Gts ) \\
     &\quad + (V^{t_s+1}_{\vx_r \vx}\fxts - \GtsT \Quuts \Kts ) + (V^{t_s+1}_{\vx_r \vx_r} - \GtsT \Quuts \Gts) \\
     &= V^{t_s}_{\vx\vx } + V^{t_s}_{\vx\vxr} + {V^{t_s\transpose}_{\vx\vxr}} + V^{t_s+1}_{\vx_r \vx_r} - \GtsT \Quuts \Gts \\
     &= V^{t_s}_{\vx\vx } + V^{t_s}_{\vx\vxr} + {V^{t_s\transpose}_{\vx\vxr}} + V^{t_f+1}_{\vx\vx} - \textstyle \sum_{t \in [t_s, t_f]} \GtT\Quut\Gt \comma  \numberthis  \label{eq:vxx-res2-app}
\end{align*}
where the last equalities in Eq.~(\ref{eq:vx-res-app}, \ref{eq:vxx-res2-app}) follow by applying the recursions 
\begin{equation}
\begin{split}
V^t_{\vx_r}    &\triangleq V^{t+1}_{\vx_r}    - \GtT\Quut\kt \comma \quad \text{ }\text{ } V^{t_f+1}_{\vxr}   =V^{t_f+1}_{\vx} \\
V^t_{\vxr\vxr} &\triangleq V^{t+1}_{\vxr\vxr} - \GtT\Quut\Gt \comma \quad V^{t_f+1}_{\vxr\vxr}=V^{t_f+1}_{\vx\vx} \period
\end{split}
\end{equation}
Thus we conclude the proof.

\subsection{Derivation in Section \ref{sec:32}} \label{app:sec32-dev}

\def\vU{{\hat{\vu}}}
\newcommand\overmat[2]{%
  \makebox[0pt][l]{$\smash{\color{white}\overbrace{\phantom{%
    \begin{matrix}#2\end{matrix}}}^{\text{\color{black}#1}}}$}#2}
\newcommand\bovermat[2]{%
  \makebox[0pt][l]{$\smash{\overbrace{\phantom{%
    \begin{matrix}#2\end{matrix}}}^{\text{#1}}}$}#2}
\newcommand\bundermat[2]{%
  \makebox[0pt][l]{$\smash{\underbrace{\phantom{%
    \begin{matrix}#2\end{matrix}}}_{\text{#1}}}$}#2}
\newcommand\partialphantom{\vphantom{\frac{\partial e_{P,M}}{\partial w_{1,1}}}}

Here we provide the derivation of Eq.~(\ref{eq:du-coop}, \ref{eq:dv-coop}).
Recall in \eq{\ref{eq:coop-traj}} the cooperative Bellman objective $Q_t(\vx_r,\vx_t,\vu_t,\vv_t)$ and 
expand it wrt all variables to the second order.
\begin{align} 
\delta Q_t =
\frac{1}{2}
\left[\begin{array}{c}
	\mathbf{1} \\ {\delta \vx_t} \\ {\delta \vx_r} \\ {\delta \vu_t} \\ {\delta \vv_t}
\end{array}\right]^{\transpose}
\left[\begin{array}{lllll}
	\mathbf{0} & {\Qxt^{\transpose}} & {\Qyt^{\transpose}} & {\Qut^{\transpose}} & {\Qvt^{\transpose}} \\ 
	{\Qxt} & {\Qxxt}& {\Qxyt} & {\Qxut} & {\Qxvt} \\ 
	{\Qyt} & {\Qyxt}& {\Qyyt} & {\Qyut} & {\Qyvt} \\ 
	{\Qut} & {\Quxt}& {\Quyt} & {\Quut} & {\Quvt} \\
	{\Qvt} & {\Qvxt}& {\Qvyt} & {\Qvut} & {\Qvvt}
\end{array}\right]
\left[\begin{array}{c}
	\mathbf{1} \\ {\delta \vx_t} \\ {\delta \vx_r} \\ {\delta \vu_t} \\ {\delta \vv_t}
\end{array}\right]
\period
\end{align}
Similar to section \ref{app:sec31-dev} where we consider the augmented state $\vX_t \triangleq [\vx_t,\vx_r]^\transpose$,
here we can additionally interpret the joint control as $\vU_t \triangleq [\vu_t,\vv_t]^\transpose$.
The derivatives of the state-control-augmented Bellman objective $\hat{Q}_t(\vX_r,\vU_t)$ thus follow
\begin{align} 
\hat{Q}^t_\vU =
\left[\begin{array}{c}
	{\Qut} \\ {\Qvt}
\end{array}\right]
\comma \quad
\hat{Q}^t_{\vU\vX} =
\left[\begin{array}{ll}
	{\Quxt} & {\Quyt} \\
	{\Qvxt} & {\Qvyt}
\end{array}\right]
\comma \quad
\hat{Q}^t_{\vU\vU} =
\left[\begin{array}{ll}
	{\Quut} & {\Quvt} \\
	{\Qvut} & {\Qvvt}
\end{array}\right]
\comma
\end{align}
and the feedback policy in this case is given by 
\begin{align*} 
\delta \vU_t^* (\delta \vX_t) &= -(\hat{Q}^t_{\vU\vU})^{\Inv} (\hat{Q}^t_\vU + \hat{Q}^t_{\vU\vX} \delta \vX_t) \\
&= 
-\left[\begin{array}{ll}
	{\Quut} & {\Quvt} \\
	{\Qvut} & {\Qvvt}
\end{array}\right]^{\Inv}
\left(
\left[\begin{array}{c}
	{\Qut} \\ {\Qvt}
\end{array}\right]
+
\left[\begin{array}{ll}
	{\Quxt} & {\Quyt} \\
	{\Qvxt} & {\Qvyt}
\end{array}\right]
 \delta \vX_t \right)
\period \numberthis \label{eq:coop-du-dev}
\end{align*}
Now, we apply the block-matrices inversion with the Schur complement by recalling
\begin{align*} 
\\
\begin{bmatrix}
	{\Quut} & {\Quvt} \\
	{\Qvut} & {\Qvvt}
\end{bmatrix}^{\Inv}
=
\begin{bmatrix}
	(\bovermat{$\QuutC$}{\Quut-\Quvt(\Qvvt)^\Inv\Qvut})^\Inv & {-(\QuutC)^\Inv\Quvt(\Qvvt)^\Inv} \\
	{-(\QvvtC)^\Inv\Qvut(\Quut)^\Inv} & (\bundermat{$\QvvtC$}{\Qvvt-\Qvut(\Quut)^\Inv\Quvt})^\Inv 
\end{bmatrix}
\period \numberthis \label{eq:schur}
\\
\end{align*}
Substitute \eq{\ref{eq:schur}} into \eq{\ref{eq:coop-du-dev}} and after some algebra, we will arrive at
\begin{align} 
\delta \vU_t^* (\delta \vX_t)
&= 
\begin{bmatrix}
	{\kut} \\
	{\Ivt}
\end{bmatrix}
+
\begin{bmatrix}
	{\Kut} & {\Gut} \\
	{\Hvt} & {\Lvt} 
\end{bmatrix}
\delta \vX_t
=
\begin{bmatrix}
	{\kut + \Kut \dvx_t + \Gut \dvx_r} \\
	{\Ivt + \Hvt \dvx_t + \Lvt \dvx_r}
\end{bmatrix}
\triangleq
\begin{bmatrix}
	{\delta \vu_t^*(\delta \vx_t,\delta \vx_r)} \\
	{\delta \vv_t^*(\delta \vx_t,\delta \vx_r)}
\end{bmatrix}
\comma
\end{align}
where 
\begin{subequations}
\begin{align}
\kut &= -(\QuutC)^\Inv (\Qut - \Quvt(\Qvvt)^\Inv\Qvt)  \comma \\
\Kut &= -(\QuutC)^\Inv (\Quxt- \Quvt(\Qvvt)^\Inv\QBt)  \comma \\
\Gut &= -(\QuutC)^\Inv (\QAt - \Quvt(\Qvvt)^\Inv\Qvyt) \comma \\
\Ivt &= -(\QvvtC)^\Inv (\Qvt - \Qvut(\Quut)^\Inv\Qut)  \comma \\
\Hvt &= -(\QvvtC)^\Inv (\QBt - \Qvut(\Quut)^\Inv\Quxt) \comma \\
\Lvt &= -(\QvvtC)^\Inv (\Qvyt - \Qvut(\Quut)^\Inv\QAt) \comma
\end{align}
\end{subequations}
which conclude Eq.~(\ref{eq:du-coop}, \ref{eq:dv-coop}).

\def\Vht{{V^{t}_{\vh}}}

\section{Derivation in Section \ref{sec:gt-ddp}} \label{app:sec4-dev}

\subsection{Preliminary on Second-Order Kronecker Factorization} \label{app:pre-kron}
Popular curvature factorization methods, such as KFAC \cite{martens2015optimizing} and EKFAC \cite{george2018fast},
rely on the fact that for feedforward networks:
\begin{align} \label{eq:dnn-dyn}
\vx_{t+1} = \sigma_t (\vh_t) \comma \quad
\vh_t \equiv \mW_t \vx_t + \vb_t \comma
\end{align}
where $\sigma_t$ is the nonlinear activation function and $\vh_t$ denotes the pre-activation vector,
we have
$J^t_{\vu} = \vx_t \otimes J^t_{\vh}$.
$\otimes$ denotes the Kronecker product and $J_t$ is the per-stage objective defined in \eq{\ref{eq:Jt}}.
Thus, the Gauss-Newton (GN) approximation of $J^t_{\vu \vu}$ can be computed as
\begin{align} \label{eq:kfac}
    J^t_{\vu \vu} \approx
    \E{[J^t_{\vu} J_{\vu}^{t \text{ } \transpose}]}
   = \E{[(\vx_t \otimes J^t_{\vh}) (\vx_t \otimes J^t_{\vh})^\transpose]}
   \approx \E{[(\vx_t \vx_t^\transpose)]} \otimes \E{[( J^t_{\vh} {J^t_{\vh}}^\transpose)]}
    \comma
\end{align}
where the expectation is taken over the mini-batch.

The factorization in \eq{\ref{eq:kfac}} is also applicable to DDP and GT-DDP, as
\eq{\ref{eq:dnn-dyn}} can be expressed by $\vx_{t+1}=f_t(\vx_t,\vu_t)$, with $\vu_t \triangleq [\vectorize(\mW_t), \vb_t]^\transpose$;
thus it is a valid dynamics.
Further, we have
\begin{align}
	\futT \Vxt = \vx_t \otimes \Vht, \quad \text{where } \Vht=\sigma_\vh^{t \text{ } \transpose}\Vxt
\end{align}
is the derivative of the value function wrt to the pre-activation.
Following similar derivation, we will arrive at the Kronecker approximation of ${Q}^t_{\vu \vu}$:
\begin{align} \label{eq:ekfac-ddp}
    {Q}^t_{\vu \vu} \approx
    \E{[Q^t_{\vu} {Q^t_{\vu}}^\transpose]}
   =\E{[(\vx_t \otimes \Vht) (\vx_t \otimes \Vht)^\transpose]}
   \approx \E{[\vx_t \vx_t^\transpose]} \otimes \E{[\Vht \Vht^\transpose]}
    \period
\end{align}
The Kronecker factorization allows us to compute the preconditioned update efficiently by
noticing that for matrices $\mA \in \mathbf{R}^{n\times n}$, $\mB \in \mathbf{R}^{m\times m}$, and $\mX \in \mathbf{R}^{m\times n}$,
we have
\begin{align}
	(\mA \otimes \mB)\vectorize (\mX) = \vectorize (\mB \mX \mA^\transpose) \comma \label{eq:kron1}
\end{align}
where $\vectorize$ denotes the vectorization.
Here, we shall interpret $\mA$ and $\mB$ respectively as $\E{[\vx_t \vx_t^\transpose]}$ and $\E{[\Vht \Vht^\transpose]}$.
Additionally, the following properties will become handy for the later derivation.
\begin{align}
	(\mA \otimes \mB)^\Inv &= \mA^\Inv \otimes \mB^\Inv \label{eq:kron2} \\
	(\mA \otimes \mB)^\transpose &= \mA^\transpose \otimes \mB^\transpose \period \label{eq:kron3}
\end{align}

\subsection{Derivation of Theorem \ref{thm:coop}} \label{app:sec41-dev}

Let us consider two distinct layers, $f(\vx_\vu,\vu)$ and $h(\vx_\vv,\vv)$,
and denote the propagation rules of their pre-activation, along with the Kronecker factorization, respectively as
\begin{equation}
\begin{split} \label{eq:dnn-dyn-coop}
\vh_\vu &= \vu \vx_\vu \comma \quad \Quu \approx \Axx \otimes \Bxx \triangleq \Auu \otimes \Buu
\comma \\
\vh_\vv &= \vv \vx_\vv \comma \quad \text{ } \Qvv \approx \Ayy \text{ } \otimes \Byy \triangleq \Avv \otimes \Bvv
\comma
\end{split}
\end{equation}
where $\vg_\vu \equiv V_{\vh_\vu}$ and $\vg_\vv \equiv V_{\vh_\vv}$ for notational simplicity.
We drop the bias in the propagation rules but note that our derivation extends to the bias cases.
Following Eq.~(\ref{eq:kron1}, \ref{eq:kron2}), the preconditioned update, take $\kt$ for instance, can be computed by
$\kt \triangleq -\QuuInv\vectorize(\Qu) \approx -\vectorize(\BuuInv\Qu\AuuInvT)$.

Now consider the CG formulation where the two layers are placed parallel in a residual network.
\ref{app:sec32-dev} suggests that one can derive
the cooperative representation by considering the joint parametrization $[\vu,\vv]^\transpose$ and
state augmentation $\vX=[\vx_\vu,\vx_\vv]^\transpose$.
To this end,
we interpret \eq{\ref{eq:dnn-dyn-coop}} as an augmented dynamics and rewrite it compactly as
\begin{align}
	\begin{bmatrix} \vh_\vu \\ \vh_\vv \end{bmatrix}
	=
	\begin{bmatrix} \vu & \mathbf{0} \\ \mathbf{0} & \vv \end{bmatrix}
	\begin{bmatrix} \vx_\vu \\ \vx_\vv \end{bmatrix}
\Leftrightarrow \hat{\vh} = \vw \hat{\vx}
\period
\end{align}
The approximated Hessian can thus be factorized as
$
Q_{\vw\vw} \approx
A_{\vw\vw} \otimes B_{\vw\vw}
$,
where
\begin{equation}
\begin{split}
	A_{\vw\vw} &=
	\E[\hat{\vx} \hat{\vx}^\transpose]
	=
		\begin{bmatrix}
			\E[{\vx_\vu} {\vx_\vu}^\transpose] && \E[{\vx_\vu} {\vx_\vv}^\transpose] \\
			\E[{\vx_\vv} {\vx_\vu}^\transpose] && \E[{\vx_\vv} {\vx_\vv}^\transpose]
		\end{bmatrix}
	=
		\begin{bmatrix}
			\Auu & \Auv \\
			\Avu & \Avv
		\end{bmatrix}
\\
	B_{\vw\vw} &=
	\E[\hat{\vg} \hat{\vg}^\transpose]
	=
		\begin{bmatrix}
			\E[{\vg_\vu} {\vg_\vu}^\transpose] && \E[{\vg_\vu} {\vg_\vv}^\transpose] \\
			\E[{\vg_\vv} {\vg_\vu}^\transpose] && \E[{\vg_\vv} {\vg_\vv}^\transpose]
		\end{bmatrix}
	\text{ }\text{ }=
		\begin{bmatrix}
			\Buu & \Buv \\
			\Bvu & \Bvv
		\end{bmatrix}
\end{split}
\end{equation}
are the Kronecker blocks.
Their inverse matrices are given by the Schur component (c.f. \eq{\ref{eq:schur}}):
\begin{equation}
\begin{split}
	A_{\vw\vw}^\Inv &=
		\begin{bmatrix}
			\AuuCInv & - \AuuCInv \Auv \AvvInv \\
			- \AvvCInv \Avu \AuuInv & \AvvCInv
		\end{bmatrix}
	\comma \quad \text{where}
	\begin{cases}
		\AuuC \triangleq \Auu - \Auv \AvvInv \Avu \\
		\AvvC \triangleq \Avv - \Auv \AvvInv \Avu
	\end{cases}
\\
	B_{\vw\vw}^\Inv &=
		\begin{bmatrix}
			\BuuCInv & - \BuuCInv \Buv \BvvInv \\
			- \BvvCInv \Bvu \BuuInv & \BvvCInv
		\end{bmatrix}
	\comma \quad \text{where}
	\begin{cases}
		\BuuC \triangleq \Buu - \Buv \BvvInv \Bvu \\
		\BvvC \triangleq \Bvv - \Buv \BvvInv \Bvu
	\end{cases}
\end{split} \label{eq:ABcoop}
\end{equation}

Now, we are ready to derive Theorem \ref{thm:coop}.
First notice that the preconditioned open gain can be computed by
\begin{equation}
\begin{split}
	-Q_{\vw\vw}^\Inv \vectorize(\begin{bmatrix} \Qu & \mathbf{0} \\ \mathbf{0} & \Qv \end{bmatrix})
=	-(A_{\vw\vw}^\Inv \otimes B_{\vw\vw}^\Inv) \vectorize(\begin{bmatrix} \Qu & \mathbf{0} \\ \mathbf{0} & \Qv \end{bmatrix})
=	-\vectorize(B_{\vw\vw}^\Inv \begin{bmatrix} \Qu & \mathbf{0} \\ \mathbf{0} & \Qv \end{bmatrix} A_{\vw\vw}^{-\transpose})
\end{split} \label{eq:46}
\end{equation}
Expanding \eq{\ref{eq:46}} by substituting $B_{\vw\vw}^\Inv$ and $A_{\vw\vw}^{-\transpose}$ with \eq{\ref{eq:ABcoop}},
after some algebra we will arrive at
\begin{equation}
\begin{split}
	\ku
&\approx	-\vectorize(\BuuCInv \Qu \AuuCInvT + \BuuCInv \Buv \BvvInv \Qv (\AuuCInv \Auv \AvvInv )^\transpose) \comma\\
	\Iv
&\approx	-\vectorize(\BvvCInv \Qv \AvvCInvT + \BvvCInv \Bvu \BuuInv \Qu (\AvvCInv \Avu \AuuInv )^\transpose) \comma
\end{split} \label{eq:47}
\end{equation}
which give \textit{the Kronecker approximation of the cooperative open gains}.
The Kronecker factorization for each cooperative matrix can be obtained by decomposed \eq{\ref{eq:47}} into the following
\begin{equation}
\begin{split}
\ku
=&	-\vectorize(\BuuCInv \Qu \AuuCInvT + \BuuCInv \Buv \BvvInv \Qv (\AuuCInv \Auv \AvvInv )^\transpose) \\
=&	-\vectorize(\BuuCInv (\Qu + \Buv \BvvInv \Qv \AvvInvT \AuvT) \AuuCInvT) \\
=&	-(\AuuCInv \otimes \BuuCInv)\vectorize(\Qu + \Buv \BvvInv \Qv \AvvInvT \AuvT) \\
=&	-(\AuuCInv \otimes \BuuCInv)(\vectorize(\Qu) + \vectorize(\Buv \BvvInv \Qv \AvvInvT \AuvT)) \\
=&	-(\AuuCInv \otimes \BuuCInv)(\vectorize(\Qu) + (\Auv \otimes \Buv) \vectorize(\BvvInv \Qv \AvvInvT)) \\
=&
-(\underbrace{\AuuCInv \otimes \BuuCInv}_{\approx \QuuCInv})(\vectorize(\Qu)
\underbrace{+ (\Auv \otimes \Buv)}_{\approx -\Quv}
  (\underbrace{\AvvInv \otimes \BvvInv}_{\approx \QvvInv}) \vectorize(\Qv))
\comma
\end{split} \label{eq:kron-dev}
\end{equation}
where we recall the definition
$\ku \triangleq -\QuuCInv(\vectorize(\Qu)-\Quv \QvvInv \vectorize(\Qv))$.
Similarly, it can be readily verified that
$\QvvCInv \approx \AvvCInv \otimes \BvvCInv$.
Note that when $\Quv$ vanishes, i.e. $\Auv=\Buv=\mathbf{0}$,
\eq{\ref{eq:kron-dev}} will degenerate to original Kronecker factorization for the non-cooperative update.
Thus, we conclude the proof.

\subsection{Derivation of Corollary \ref{coro:5}} \label{app:coro5-dev}

Before deriving Corollary \ref{coro:5}, we first review the eigen-basis representation of the Kronecker approximation appeared in \citet{george2018fast}. Recall the factorization $\Quu \approx \Auu \otimes \Buu$ and let
$\Auu = \mU_{A} \Sigma_A \mU_{A}^\transpose$,
$\Buu = \mU_{B} \Sigma_B \mU_{B}^\transpose$
be their eigen-decomposition.
We can rewrite the Kronecker factorization in its eigen-basis
\begin{equation}
\begin{split}
	\Auu \otimes \Buu
=&	(\mU_{A} \Sigma_A \mU_{A}^\transpose) \otimes (\mU_{B} \Sigma_B \mU_{B}^\transpose) \\
=&	(\mU_{A} \otimes \mU_{B}) (\Sigma_A \otimes \Sigma_B) (\mU_{A} \otimes \mU_{B})^\transpose \\
\triangleq& \text{ } \mU \Sigma_{\vu\vu} \mU^\transpose \comma
\end{split} \label{eq:49}
\end{equation}
where $\mU$ is the eigen-basis of the Kronecker factorization. $\Sigma_{\vu\vu}\triangleq\diag({\lambda}_{\vu\vu})$ contains eigenvalues along the diagonal entries.
In practice, we will also add a positive Tikhonov coefficient $\gamma>0$ for regularization purpose.

Now, observe that for the cooperative game module in Fig.~\ref{fig:cg-net}, we have
\begin{align}
	\Auu = \Auv = \Avv \comma \quad \Buu = \Buv = \Bvv \comma
\end{align}
since the two layers share the same input $\vx_\vu=\vx_\vv$ and output derivative $\vg_\vu=\vg_\vv$.
In other words,
$\Quv$ and $\Qvv$ are factorized by the same Kronecker blocks with $\Quu$;
thus they share the same eigen-basis $\mU$. %
The cooperative matrix $\QuuC$ can thus be rewritten as
\begin{equation}
\begin{split}
	\QuuC
	&= \Quu - \Quv \QvvInv \QuvT \\
	&= \Auu \otimes \Buu - (-\Auv \otimes \Buv) (\Avv \otimes \Bvv)^\Inv (-\Auv \otimes \Buv)^\transpose \\
	&= \mU (\gamma\mI+\Sigma_{\vu\vu}) \mU^\transpose - (-\mU \Sigma_{\vu\vu} \mU^\transpose) (\mU (\gamma\mI+\Sigma_{\vu\vu})^\Inv \mU^\transpose) (-\mU \Sigma_{\vu\vu} \mU^\transpose)^\transpose \\
	&=  \mU \tilde{\Sigma}_{\vu\vu} \mU^\transpose \comma
\end{split}
\end{equation}
where $\tilde{\Sigma}_{\vu\vu} = \gamma\mI + \diag(\tilde{\lambda}_{\vu\vu})$ and
\begin{align}
	\tilde{\lambda}_{\vu\vu}^i
	= \lambda_{\vu\vu}^i - \frac{(\lambda_{\vu\vu}^i)^2}{\gamma + \lambda_{\vu\vu}^i}
	= \frac{\gamma}{\gamma+{\lambda}_{\vu\vu}^i} {\lambda}_{\vu\vu}^i \period
\end{align}
In short,
the cooperative matrix $\QuuC$ admits a scaling in the eigen-basis of its non-cooperative variant.

\subsection{Proof for Proposition \ref{prop:bp2ddp2}} \label{app:prop4-dev}

Recall the connection we made in Sec. \ref{sec:ocp-dnn} and \ref{sec:ddp}.
It is sufficient to show that when the two conditions in Proposition \ref{prop:bp2ddp2} are met,
we will have Eq.~(\ref{eq:du-star-ddp}, \ref{eq:Qt}) collapse exactly with Eq.~(\ref{eq:du-star}, \ref{eq:Jt}).
First, notice that at the final layer, we have
$V^T_{\vx}=J^T_{\vx}=\nabla_{\vx}\phi$ and
$V^T_{\vx\vx}=J^T_{\vx\vx}=\nabla_{\vx}^2\phi$  without any condition.
Further, Eq.~(\ref{eq:vx_res_app}, \ref{eq:vxx_res_app})
suggest that when all mixed partial derivatives between parameter and activation vanish,
the backward dynamics of ($V^t_\vx$,$V^t_{\vx\vx}$) degenerates to ($\Qxt,\Qxxt$).
The derivatives of $J_t$ wrt $\vx_t$ in this case (c.f. \eq{\ref{eq:Jt}}),
\begin{align*}
	J^t_{\vx} = \fxtT J^{t+1}_{\vx} \comma \quad
	J^t_{\vx\vx} = \fxtT J^{t+1}_{\vx\vx} \fxt \comma
\end{align*}
are the same as the backward dynamics for ($V^t_\vx$,$V^t_{\vx\vx}$),
\begin{align*}
	V^t_\vx = \Qxt = \fxtT \Vxt \comma \quad
	V^t_{\vx\vx} = \Qxxt = \fxtT \Vxxt \fxt \period
\end{align*}
Thus the two functionals $J_t$ and $V_t$ coincide with each other.

Next,
when the parameters between distinct layers are uncorrelated,
we will have $\Quvt=\Qvut=\mathbf{0}$ at all stages.
The cooperative precondition matrices, if exist along the network, degenerate to the curvature approximation it uses to approximate the parameter Hessian.
In fact, we will have
\begin{align*}
	\Qut = J^t_{\vu} \comma \quad
	\QuutC = \Quut = J^t_{\vu\vu} \period
\end{align*}
Thus, the update rule \eq{\ref{eq:du-star-ddp}} also collapses to \eq{\ref{eq:du-star}}.

\section{Experiment Detail} \label{app:exp}

\subsection{Experiment Setup in Section \ref{sec:experiment} and \ref{sec:discussion}} \label{app:exp-set}
Network architectures for classification task are shown in Fig.~\ref{fig:architecture}.
We use $1$ residual block for DIGITS, MNIST, SVHN dataset and $4$ residual blocks for CIFAR-10.
For CIFAR-100, we use ResNet18 \cite{he2016deep} architecture.
All networks use ReLU activation for the intermediate layers and identity mapping at the last prediction layer.
The batch size is set to $128$ for all data set except $8$ for DIGITS.
As for section \ref{sec:discussion},
the network contains $3$ convolution CGBs (c.f. Fig.~\ref{fig:cg-net}), $1$ fully-connected CGB, and finally $1$ standard fully-connected layer with identity mapping.
We use Tanh activation for this experiment but note that similar trend can be observed for ReLU.
The batch size is set to $12$.
Regarding the machine information,
we conduct our experiments on GTX 1080 TI, RTX TITAN, four Tesla V100 SXM2 16GB on AWS, and eight GTX TITAN X.
All experiments are implemented and conducted with Pytorch \cite{paszke2017automatic}.
We use the implementation in \url{https://github.com/Thrandis/EKFAC-pytorch} for EKFAC baseline.

\subsection{Additional Result and Discussion} \label{app:add-exp}

\begin{figure}[h]
  \begin{center}
  \includegraphics[width=0.35\textwidth]{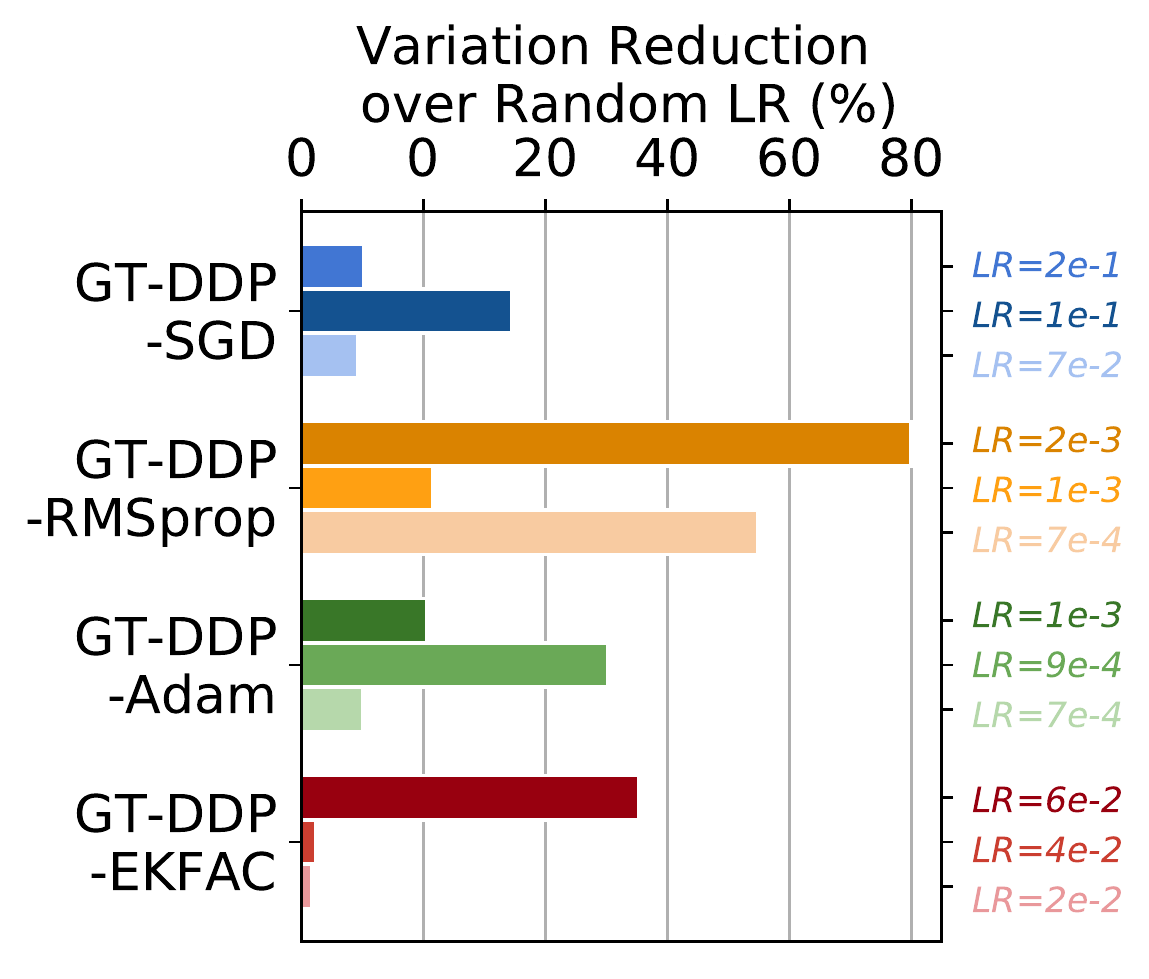}
  \end{center}
  \caption{
  Variation reduction over $3$ different learning rates for each GT-DDP variant on CIFAR-10.
  We report the value $(\Var_{\text{GT-DDP-Baseline}}-\Var_{\text{Baseline}})/\Var_{\text{Baseline}}$,
  where each variance is computed over $3$ random seeds.
  }
  \label{fig:var-lr}
  \vskip -0.1in
\end{figure}

\textbf{Variation Reduction Over Different Learning Rate.}
Recall
Fig.~\ref{fig:cifar-100-var} reports the variation reduction on the hyper-parameter used in Table \ref{table:training}.
Here we provide additional results and show that the robustness gained from GT-DDP integration remains consistent across different hyper-parameters.
Particularly,
in Fig.~\ref{fig:var-lr} we report the variance difference on $3$ different learning rates for each GT-DDP variant.
We use the same setup as in Fig. \ref{sec:experiment}, i.e.
we keep all hyper-parameters the same for each experiment so that the performance difference only comes from the existence of feedback policies.
For all cases, having additional updates from GT-DDP stabilizes the training dynamics by reducing its variation over random initialization.

\end{document}